\documentclass[runningheads]{llncs}

\usepackage{eccv}

\usepackage{eccvabbrv}

\usepackage{amsmath}
\usepackage{xfrac}
\usepackage{multirow}
\usepackage{graphicx}
\usepackage{booktabs}
\usepackage{makecell}

\DeclareMathOperator*{\argmax}{argmax}

\usepackage[accsupp]{axessibility}  

\usepackage{hyperref}

\usepackage{selectp}

\usepackage{orcidlink}

\begin{document}

\title{Embedding Geometries of Contrastive Language-Image Pre-Training} 

\author{Jason Chuan-Chih Chou\inst{1} \and
Nahid Alam\inst{1,2}}

\authorrunning{J.~Chou and N.~Alam}

\institute{
Cohere For AI Community\\
\email{\{chuanchih,nahid.m.alam\}@gmail.com} \and
Cisco Meraki\footnote{Work does not relate to position at Cisco Meraki.}
}

\maketitle

\begin{abstract}
  Since the publication of CLIP, the approach of using InfoNCE loss for contrastive pre-training has become widely popular for bridging two or more modalities. Despite its wide adoption, CLIP's original design choices of L2 normalization and cosine similarity logit have rarely been revisited. We have systematically experimented with alternative geometries and softmax logits for language-image pre-training and identified that variants with intuitive Euclidean geometry, Euclidean CLIP (EuCLIP), match or exceed the performance of CLIP and support hierarchical relationships at least as well as more complicated hyperbolic alternative.
  \keywords{CLIP \and Euclidean \and hyperbolic}
\end{abstract}

\section{Introduction}
\label{sec:intro}

Originally proposed as ConVIRT for the application of medical imaging~\cite{DBLP:conf/mlhc/0004JMML22} but scaled up and publicized as CLIP~\cite{DBLP:conf/icml/RadfordKHRGASAM21}, contrastive pre-training using massive image-text pairs with InfoNCE loss enables models to perform zero-shot image classification and retrieval, without the need of manual labels of predefined categories. Furthermore, since such pre-training only requires encoders of respective modalities without any specific cross-modal modelling, it has been applied to modalities beyond image and text like audio and video and culminated in the 6-modality model of ImageBind~\cite{DBLP:conf/cvpr/GirdharELSAJM23}. In contrast to its wide applicability, the original design choices of CLIP have largely stayed the same, namely L2-normalizing the embeddings and using cosine similarity as the softmax logit. Desai \etal~\cite{DBLP:conf/icml/DesaiNR0V23} proposed MERU, which exponentially lifts the embeddings onto the Lorentz hyperboloid instead of L2 normalization. As a standard model of hyperbolic geometry, the Lorentz hyperboloid enables MERU to use negative Lorentzian distance as the softmax logit and use hyperbolic entailment loss to enforce hierarchical relationships between paired text and images. Curiously, they identified the embedding space of CLIP as Euclidean, even though L2 normalization puts all the $n$-dim CLIP embeddings on the $(n-1)$-sphere $S^{n - 1} = \left\{ \mathbf{x} \in \mathbb{R}^n : \left\| \mathbf{x} \right\| = 1 \right\}$, one of the standard models of elliptic geometry. On the other hand, cosine similarity of the CLIP model can't be considered negative of a distance metric\footnote{In order for $d(\mathbf{x}, \mathbf{x}) = 0$, the potential distance metric must be proportional to $1 - \mathbf{x} \cdot \mathbf{y}$, sometimes called ``cosine distance''. However, it doesn't satisfy triangle inequality.}, leaving the question of whether the design choice of softmax logit is optimal open for either geometry.

In pursue of these open questions, we have systematically tested various embedding geometries of contrastive language-image pre-training models in combination with alternative softmax logit, with emphasis on the unexplored Euclidean geometry. We find that
\begin{itemize}
\item It makes no difference with elliptic geometry whether the softmax logit is cosine similarity (CLIP) or negative geodesic arccos distance; 
\item For both Euclidean and hyperbolic geometries, the final LayerNorm of the vision and text transformers degrades performance and negative distance squared logit outperforms negative distance logit possibly due to the implicit L2 regularization;
\item Euclidean CLIP (EuCLIP) matches or exceeds the performance of CLIP and supports hierarchical relationships at least as well as the more complicated MERU.
\end{itemize}

\section{Related Work}

\subsection{Alternative Loss for Language-Image Pre-training}
Alternative pre-training objectives have been proposed for language-image models, including CoCa\cite{DBLP:journals/tmlr/YuWVYSW22}, OTTER\cite{DBLP:conf/iclr/WuCZGGV22}, and SigLIP\cite{DBLP:conf/iccv/ZhaiM0B23}. CoCa\cite{DBLP:journals/tmlr/YuWVYSW22} still uses InfoNCE loss but with the addition of captioning loss by a multimodal text decoder. OTTER\cite{DBLP:conf/iclr/WuCZGGV22} deviates further from InfoNCE loss by taking text-text and image-image similarities into account and targeting modified matching probabilities that no longer form an identity matrix. Finally, the most recent SigLIP\cite{DBLP:conf/iccv/ZhaiM0B23} effectively runs logistic regression on all positive and negative text-image pairs instead of contrastive loss. Other than MERU~\cite{DBLP:conf/icml/DesaiNR0V23} however, alternative softmax logit is less explored.

\subsection{Hyperbolic vs. Euclidean Geometries}
Nickel-Kiela~\cite{DBLP:journals/corr/NickelK17} first proposed using hyperbolic embeddings trained with InfoNCE loss to predict hierarchical relations in the WordNet nouns hypernymy tree and compared their performance to that of Euclidean embeddings. Their conclusion, however, is challenged by Bansal-Benton\cite{DBLP:journals/corr/abs-2109-07488} who pointed out that Euclidean embeddings become competitive when unnecessary constraint on their norm is removed. Ganea \etal~\cite{DBLP:conf/icml/GaneaBH18} proposed entailment loss as an alternative to InfoNCE for the same task and dataset and similarly compared the performance of embeddings in hyperbolic vs. Euclidean geometries. In the field of reinforcement learning, Cetin \etal~\cite{DBLP:conf/iclr/CetinCBH23} compared the performance of PPO\cite{DBLP:journals/corr/SchulmanWDRK17} agents using hyperbolic embeddings vs. the ones using Euclidean embeddings to represent states. Whereas in the field of computer vision, Khrulkov \etal~\cite{DBLP:conf/cvpr/KhrulkovMUOL20} proposed Hyperbolic ProtoNet for classification using prototype embeddings and compared its performance on few-shot classification to the original Euclidean ProtoNet\cite{DBLP:conf/nips/SnellSZ17}.

\subsection{Layer Normalization of Transformers}
The role of Layer Normalization (LN) in the transformer architecture has received much scrutiny\cite{DBLP:conf/icml/XiongYHZZXZLWL20,DBLP:conf/acl/Brody0Y23}. When transformer was first proposed, LN is placed between the residual blocks (Post-LN Transformer) \cite{DBLP:conf/nips/VaswaniSPUJGKP17} but later a variant in which LN is placed within the residual blocks is proposed as an alternative (Pre-LN Transformer). At first, LN is also absent after the final layer\cite{DBLP:conf/iclr/BaevskiA19,ott2019fairseq} of the Pre-LN Transformer, but later most of the Pre-LN Transformer architectures including the vision transformer (ViT) have added an additional final LN\cite{DBLP:journals/corr/abs-1904-10509,DBLP:conf/iclr/DosovitskiyB0WZ21}, a change that has been much less examined.

\section{Pre-Training Loss for Language-Image Model}

For language-image pre-training, we have a dataset of text-image pairs that we divide into mini-batches $\mathcal{B}=\{(T_1, I_1), (T_2, I_2), \dots\}$ from which we want the model to learn representations of text and images. In this paper, we consider the following pre-training losses.

\subsection{Contrastive Loss}

One option of pre-training objective is for the model to learn to match an image from the mini-batch to its corresponding text, and vice versa. Assumed that we have an text encoder $f(\cdot)$ and a image encoder $g(\cdot)$, we can in turn consider the image and the text as the ``context'' and apply InfoNCE\cite{DBLP:journals/corr/abs-1807-03748} twice to obtain the contrastive loss $\mathcal{L}_{cont}$:

\begin{equation*}
\mathcal{L}_{cont} = -\frac 1 {2|\mathcal{B}|} \sum_{i=1}^{|\mathcal{B}|} \left(
\overbrace{\log \frac {e^{\beta \mathrm{sim}(f(T_i), g(I_i))}} {\sum_{j=1}^{|\mathcal{B}|} e^{\beta \mathrm{sim}(f(T_j), g(I_i))}}}^\text{image$\rightarrow$text softmax}
+
\overbrace{\log \frac {e^{\beta \mathrm{sim}(f(T_i), g(I_i))}} {\sum_{j=1}^{|\mathcal{B}|} e^{\beta \mathrm{sim}(f(T_i), g(I_j))}}}^\text{text$\rightarrow$image softmax}
\right)
\end{equation*}

Where $\mathrm{sim}(\cdot, \cdot)$ is some similarity function, $\beta$ is the logit scale sometimes called thermodynamic beta or inverse temperature borrowing from the physics terminology, and the density ratio is assumed to be of the form $f(T, I) = e^{\beta \mathrm{sim}(f(T), g(I))}$. The similarity function in turn depends on the underlying geometries of the model:

\subsubsection{CLIP}

If the similarity function is cosine similarity, \ie $\mathrm{sim}(\mathbf{x}, \mathbf{y}) = \frac{\mathbf{x} \cdot \mathbf{y}}{\|\mathbf{x}\| \|\mathbf{y}\|}$ where $\| \cdot \|$ is the L2 norm, we recover the loss function of CLIP. Its connection to the underlying geometry is less direct however, since cosine similarity is merely a decreasing function with respect to the geodesic arccos distance on $S^{n - 1}$. For convenience, we call the underlying geometry of CLIP models ``CLIP'' geometry.

\subsubsection{Elliptic}

We can use negative geodesic distance on $S^{n - 1}$ as the similarity function instead, which is simply $\mathrm{sim}(\mathbf{x}, \mathbf{y}) = -\arccos(\frac{\mathbf{x} \cdot \mathbf{y}}{\|\mathbf{x}\| \|\mathbf{y}\|})$. It maps pairs of $n$-dim embeddings to $[-\pi, 0]$ instead of $[-1, 1]$ and weighs similarity between them differently but preserves the ordering of cosine similarity and otherwise functions the same.

\subsubsection{Euclidean}

Euclidean geometry is the most intuitive but curiously unexplored geometry for contrastive pre-training, where the distance is simply $\|\mathbf{x} - \mathbf{y}\|$. In order to reduce dependency of the expected value on the embedding dimension $n$ and make the logit scale $\beta$ more comparable across geometries, we scale the embeddings by $\frac{1}{\sqrt{n}}$ and use $\mathrm{sim}(\mathbf{x}, \mathbf{y}) = -\frac{1}{\sqrt{n}} \|\mathbf{x} - \mathbf{y}\|$ as the similarity function. We also consider using the negative Euclidean distance squared, $\mathrm{sim}(\mathbf{x}, \mathbf{y}) = -\frac{1}{n} \|\mathbf{x} - \mathbf{y}\|^{2}$, as the similarity function for two reasons: 1. In order to calculate the Euclidean distance, we first calculate distance squared\cite{kim2021lipschitz} and then take square root, $\|\mathbf{x} - \mathbf{y}\| = \sqrt{\|\mathbf{x}\|^2 - 2 \mathbf{x} \cdot \mathbf{y} + \|\mathbf{y}\|^2}$, whose gradient blows up at the origin. 2. Distance squared logit results in a L2 regularization term on the positive pair, $\|f(T_i) - g(I_i)\|^2$, which we speculate may lead to better L2 norm distribution and representation. Furthermore, if we consider embeddings of one modality as the ``prototypes'', the distance squared logit can be reinterpreted as a linear model\cite{DBLP:conf/nips/SnellSZ17}.

\subsubsection{Hyperbolic}

For hyperbolic geometry we follow the formulation and the hyperboloid model of MERU\cite{DBLP:conf/icml/DesaiNR0V23} whose similarity function is parameterized by 3 trainable scalars: text embedding scale $\alpha_{txt}$, image embedding scale $\alpha_{img}$, and curvature parameter $c$. We first scale the encoder output by the first two scalars:
\begin{align*} 
\mathbf{u} &=  \alpha_{txt}f(T) \\ 
\mathbf{v} &=  \alpha_{img}g(I)
\end{align*}
where both $\alpha_{txt}$ and $\alpha_{img}$ are initialized to $\frac{1}{\sqrt{n}}$ for the same reason as their Euclidean counterpart. Then we use the curvature parameter $c$ to construct the exponential map at the origin $\mathbf{O}$:
\begin{equation*}
    \text{expm}_{\mathbf{O}, space}(\mathbf{u}) = \frac{\sinh(\sqrt{c} \; \lVert \mathbf{u} \rVert)}{\sqrt{c} \; \lVert \mathbf{u} \rVert} \mathbf{u}
\end{equation*}
To lift the embeddings
\begin{align*} 
\mathbf{x}_{space} &=  \text{expm}_{\mathbf{O}, space}(\mathbf{u}) \\ 
\mathbf{y}_{space} &=  \text{expm}_{\mathbf{O}, space}(\mathbf{v})
\end{align*}
To the Lorentz hyperboloid
\begin{equation*}
    \mathcal{L}^n = \{\mathbf{x} \in \mathbb{R}^{n+1} : \langle \mathbf{x},\mathbf{x} \rangle_\mathcal{L} = \sfrac{-1}{c} \} \; , \; c > 0
\end{equation*}
where $\langle \cdot , \cdot \rangle_\mathcal{L}$ is the Lorentzian inner product
\begin{equation*}
    \langle \mathbf{x},\mathbf{y} \rangle_\mathcal{L} = \mathbf{x}_{space} \cdot \mathbf{y}_{space} - x_{time} \; y_{time}
\end{equation*}
where we borrow the terminology of space dimensions and time dimension from Minkowski spacetime. The time dimension of $\mathbf{x}$ and $\mathbf{y}$ can then be inferred as
\begin{align*}
x_{time} &= \sqrt{\sfrac{1}{c} + \lVert \mathbf{x}_{space} \rVert^2} \\
y_{time} &= \sqrt{\sfrac{1}{c} + \lVert \mathbf{y}_{space} \rVert^2}
\end{align*}
Finally, the similarity function is the negative Lorentzian distance between $\mathbf{x}$ and $\mathbf{y}$:
\begin{equation*}
    \mathrm{sim}(f(T), g(I)) = -d_\mathcal{L} (\mathbf{x}, \mathbf{y}) = -\sqrt{\sfrac{1}{c}} \cdot \cosh^{-1} ( -c \; \langle \mathbf{x},\mathbf{y} \rangle_\mathcal{L} )
\end{equation*}

While the case here is less intuitive, we also explore using the negative Lorentzian distance squared as the similarity function instead: $\mathrm{sim}(f(T), g(I)) = -d_\mathcal{L} (\mathbf{x}, \mathbf{y})^2$.

\subsection{Entailment Loss}
Entailment loss, in terms of how far the embedding of the more specific concept $\mathbf{y}$ deviates from an entailment cone centered around the embedding of the more generic concept $\mathbf{x}$, was first proposed to model hierarchical concepts of WordNet\cite{DBLP:conf/icml/GaneaBH18} and has the desirable property of transitivity, \ie if $\mathbf{x}$ entails $\mathbf{y}$ and $\mathbf{y}$ entails $\mathbf{z}$, $\mathbf{x}$ entails $\mathbf{z}$. With the insight that images tend to be more specific than text, Desai \etal~\cite{DBLP:conf/icml/DesaiNR0V23} incorporated entailment loss in MERU to enforce such relationship.

\subsubsection{Euclidean}
\begin{figure}[t]
  \centering
   \includegraphics[width=0.5\linewidth]{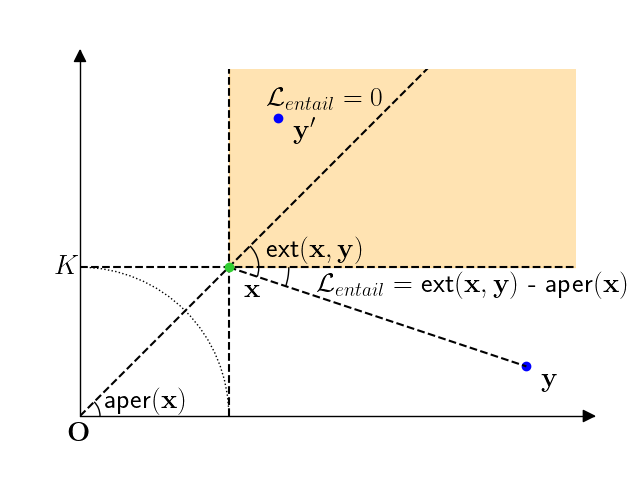}
   \caption{Euclidean entailment loss in $\mathbb{R}^2$, where $\mathbf{O}$ is the origin and $K$ is the minimum radius. For $\mathbf{x}$ on line $y = K$ its half-aperture $\text{aper}(\mathbf{x}) = \sin^{-1}(K/\lVert \mathbf{x} \rVert)$ is equal to the angle between line $\mathbf{O} \mathbf{x}$ and the x-axis. $y = K$ therefore forms one side of the entailment cone and by symmetry the entailment cone for $\mathbf{x} = (K, K)$ is simply a shifted quadrant. For $\mathbf{y}$ out of the entailment cone the entailment loss is $\text{ext}(\mathbf{x}, \mathbf{y}) - \text{aper}(\mathbf{x}), \text{ext}(\mathbf{x}, \mathbf{y}) = \pi - \angle \mathbf{O} \mathbf{x} \mathbf{y}$. For $\mathbf{y'}$ within the entailment cone the entailment loss is zero.}
   \label{fig:entailment_loss}
\end{figure}

Euclidean entailment loss was introduced in\cite{DBLP:conf/icml/GaneaBH18} and we will present a modified formulation here. Given embedding $\mathbf{x}$ in Euclidean geometry, its entailment cone is determined by the half-aperture

\begin{equation*}
    \text{aper}(\mathbf{x}) = \sin^{-1} \left( \frac{K}{\lVert \mathbf{x} \rVert} \right) \; , \; \lVert \mathbf{x} \rVert \ge K
\end{equation*}
where minimum radius $K$ is a hyperparameter. This, however, leaves entailment cone undefined for $\mathbf{x}$ with $\lVert \mathbf{x} \rVert < K$. For our implementation, we clamp $\frac{K}{\lVert \mathbf{x} \rVert}$ to make sure that it is defined everywhere in $\mathbb{R}^n$:
\begin{equation*}
    \text{aper}(\mathbf{x}) = \sin^{-1} \left(\min \left(1, \; \frac{K}{\lVert \mathbf{x} \rVert} \right) \right)
\end{equation*}
Transitivity though may not hold for $\mathbf{x}$ with $\lVert \mathbf{x} \rVert < K$. The exterior angle given by the origin $\mathbf{O}$, $\mathbf{x}$, and $\mathbf{y}$ is then

\begin{equation*}
\text{ext}(\mathbf{x}, \mathbf{y}) = \pi - \angle \mathbf{O} \mathbf{x} \mathbf{y} = \cos^{-1} \left( \frac{(\mathbf{y} - \mathbf{x}) \cdot \mathbf{x}}{\|\mathbf{y - x}\| \|\mathbf{x}\|} \right)
\end{equation*}
The entailment loss $\mathcal{L}_{entail}$ is then given by how much further $\text{ext}(\mathbf{x}, \mathbf{y})$ lies outside of the entailment cone (Figure \ref{fig:entailment_loss}):

\begin{equation*}
    \mathcal{L}_{entail}(\mathbf{x}, \mathbf{y}) = \max(0, \; \text{ext}(\mathbf{x}, \mathbf{y}) - \text{aper}(\mathbf{x}))
\end{equation*}

\subsubsection{Hyperbolic}
\label{sec:hyperbolic_entailment_loss}
For hyperbolic entailment loss, we again follow the formulation of MERU\cite{DBLP:conf/icml/DesaiNR0V23}. In its hyperboloid model, the half-aperture is given by

\begin{equation*}
    \text{aper}(\mathbf{x}) = \sin^{-1} \left( \frac{2K}{\sqrt{c} \; \lVert \mathbf{x}_{space} \rVert} \right), \; \lVert \mathbf{x}_{space} \rVert \ge \frac{2K}{\sqrt{c}}
\end{equation*}
Similar to the Euclidean counterpart, both our implementation and that of Desai \etal\footnote{In fact, the implementation of Desai \etal tends to employ more aggressive numerical smoothing, including clamping $\frac{2K}{\sqrt{c} \; \lVert \mathbf{x}_{space} \rVert}$ to $1 - \epsilon, \; \epsilon = 10^{-8}$ here. We find such numerical smoothing unnecessary for stability.} allow the half-aperture to be defined for all $\mathbf{x}_{space} \in \mathbb{R}^n$ by clamping $\frac{2K}{\sqrt{c} \; \lVert \mathbf{x}_{space} \rVert}$ to $1$:
\begin{equation*}
    \text{aper}(\mathbf{x}) = \sin^{-1} \left(\min \left(1, \; \frac{2K}{\sqrt{c} \; \lVert \mathbf{x}_{space} \rVert} \right) \right)
\end{equation*}
The exterior angle given by the origin $\mathbf{O}$, $\mathbf{x}$, and $\mathbf{y}$ is then
\begin{equation*}
    \def\cvl{c \; \langle \mathbf{x},\mathbf{y} \rangle _\mathcal{L}}
    \text{ext}(\mathbf{x}, \mathbf{y}) = \pi - \angle \mathbf{O} \mathbf{x} \mathbf{y} = \cos^{-1} \left( \frac{y_{time} + x_{time} \; \cvl{}}{\lVert \mathbf{x}_{space} \rVert \sqrt{\left( \cvl \right)^2 - 1}} \right)
\end{equation*}
The entailment loss $\mathcal{L}_{entail}$ is the same as the Euclidean counterpart in terms of $\text{aper}(\mathbf{x})$ and $\text{ext}(\mathbf{x}, \mathbf{y})$:
\begin{equation*}
    \mathcal{L}_{entail}(\mathbf{x}, \mathbf{y}) = \max(0, \; \text{ext}(\mathbf{x}, \mathbf{y}) - \text{aper}(\mathbf{x}))
\end{equation*}

For both Euclidean and hyperbolic geometry models, the total loss is then $\mathcal{L}_{cont} + \lambda \mathcal{L}_{entail}$ where the entailment loss weight $\lambda$ is another hyperparameter.

\section{Experimental Setup}

\subsection{Code}
All the experiments are conducted with PyTorch 2.0+\cite{Paszke_PyTorch_An_Imperative_2019}, modified OpenCLIP\cite{ilharco_gabriel_2021_5143773} from version v2.20.0 to v2.24.0 as we incorporated its upstream bugfixes and features, modified DataComp\cite{DBLP:conf/nips/GadreIFHSNMWGZO23} and its dependency CLIP\_benchmark (Supplementary Material \ref{sec:source_code}). In particular, we rely on its implementation of ViT-B/32 and ViT-B/16. Other than the final LN of the Pre-LN Transformer, we keep the text and image encoders unmodified including parameter initialization.

\subsection{Data}
We first tested the code and narrowed down the range of hyperparameters by training models with approximately the first 1M text-image pairs of RedCaps v1.0\cite{desai2021redcaps} but as our main experiments all the models presented here are trained with the small and medium scale datasets of DataComp\cite{DBLP:conf/nips/GadreIFHSNMWGZO23}. At each scale, we use the filtering method shown to result in the best zero-shot performance in the DataComp filtering track: Namely, CLIP score (L/14 30\%) (the top 30\% of the examples based on the OpenAI CLIP ViT-L/14 score) for the small scale and Image-based $\cap$ CLIP score (L/14 30\%) (intersection between the CLIP score filtering and filtering for the examples whose images cluster around the ImageNet classes) for the medium scale. In the Oct. 2023 - Nov. 2023 period we were able to download 87.3\% of the images of the filtered small scale dataset and 88.4\% of the images of the filtered medium scale dataset, to which we attribute the discrepancy between performance reported by the DataComp paper and our reproduction of the CLIP geometry models. We also adopt the zero-shot evaluation protocol of DataComp and use its evaluation code modified to support different embedding geometries.

\subsection{Hardware}
For the pilot tests with the 1M RedCaps slice we used a single GeForce RTX 3080 16 GB Laptop GPU while for the main experiments we use a 8 $\times$ V100 32 GB workstation. On such workstation batch size 4096 requires batch size 512 per GPU but neither ViT-B/32 nor ViT-B/16 fits with such large batch, so we use OpenCLIP's implementation of gradient accumulation to run with batch size 256 per GPU $\times$ 2 gradient accumulation steps per update to the same effect, at the price of one extra forward pass.

\subsection{Hyperparameters}
We adopt the total train compute, learning rate schedule, and optimizer configuration of the filtering track of DataComp unmodified, \ie 12.8M/128M total samples seen for the small/medium scale with maximum learning rate 5e-4, AdamW optimizer with $\beta_2 = 0.98$, cosine learning rate schedule with 500 warm-up steps, and batch size 4096. In order to make sure that these scalar hyperparameters stay positive, logit scale $\beta$, curvature parameter $c$, and text/image embedding scales $\alpha_{txt}$/$\alpha_{img}$ are all parameterized on the logarithmic scale, \eg the logit scale $\beta$ is computed as $\beta = \exp(t)$ during the forward pass on the fly with $t$ initialized as $t = \log(\frac{1}{0.07})$ for distance $d$ logit models including CLIP.

\subsubsection{Distance $d$ logit models} For models that use negative distance $d$ as softmax logit including CLIP geometry, elliptic geometry, and the distance $d$ variants of Euclidean and hyperbolic geometries, we follow the practice of\cite{DBLP:conf/cvpr/WuXYL18,DBLP:conf/icml/RadfordKHRGASAM21,DBLP:conf/icml/DesaiNR0V23} and initialize logit scale with $\beta = \frac{1}{0.07}$ as described above and clamp
it to a maximum value of 100. We also keep the rest of the hyperparameters for the distance $d$ variant of hyperbolic geometry models unmodified from MERU: curvature parameter is initialized with $c = 1$ and clamped to $[0.1, 10.0]$, minimum radius $K$ set to constant $K = 0.1$ and entailment loss weight $\lambda$ set to constant $\lambda = 0.2$ for the entailment experiments.

\subsubsection{Distance squared $d^2$ logit models} Due to the quadratic dependence on the distance, training the models that use negative distance squared $d^2$ as softmax logit with the same initial logit scale $\beta$ results in instability. Experimentally, we find that Euclidean and hyperbolic medium scale models with $d^2$ softmax logit and entailment loss may need to have initial logit scale $\beta$ lowered to $\exp(-1)$ while the rest stay stable with initial logit scale $\beta = 1$. We believe that one factor is that entailment loss significantly drives text and image embeddings apart, as we will see later. Another factor is that at the DataComp small scale we only train for 12.8M / 4096 = 3125 steps while 500 of them are warm-up steps, followed by cosine learning rate schedule. We hypothesize that with such learning rate schedule, the model is less likely to reach the combination of large embedding distance and high learning rate to become unstable. Possibly for the same reason, the model's performance is sensitive to such difference in initial logit scale at the small scale, but less so at the medium scale as long as the training is stable. Perhaps it's also worth noticing that the DataComp small scale is the only scale at which filtering method CLIP score (L/14 30\%) results in the best performance, again hinting at the qualitative difference. Finally, for the entailment experiments, we find that minimum radius $K$ set to constant $K = 0.3$ and entailment loss weight $\lambda \in [0, 0.2]$ result in stable training.

\section{Results and Discussion}
Due to the qualitative difference between the DataComp small and medium scale, we can't use the small scale to tune the hyperparameters. We therefore tune the hyperparameters at the medium scale with ViT-B/32 and then scale the best model for ImageNet, EuCLIP (Euclidean geometry, $d^2$ logit, no final LN, $K = 0.3$, $\lambda = 0.1$), up to ViT-B/16 for head-to-head comparison with CLIP and MERU (hyperbolic geometry, $d$ logit, final LN, $K = 0.1$, $\lambda = 0.2$):

\subsection{ViT-B/16 Model Head-to-head Comparison}
\begin{figure}
    \centering
    \includegraphics[width=1.0\linewidth]{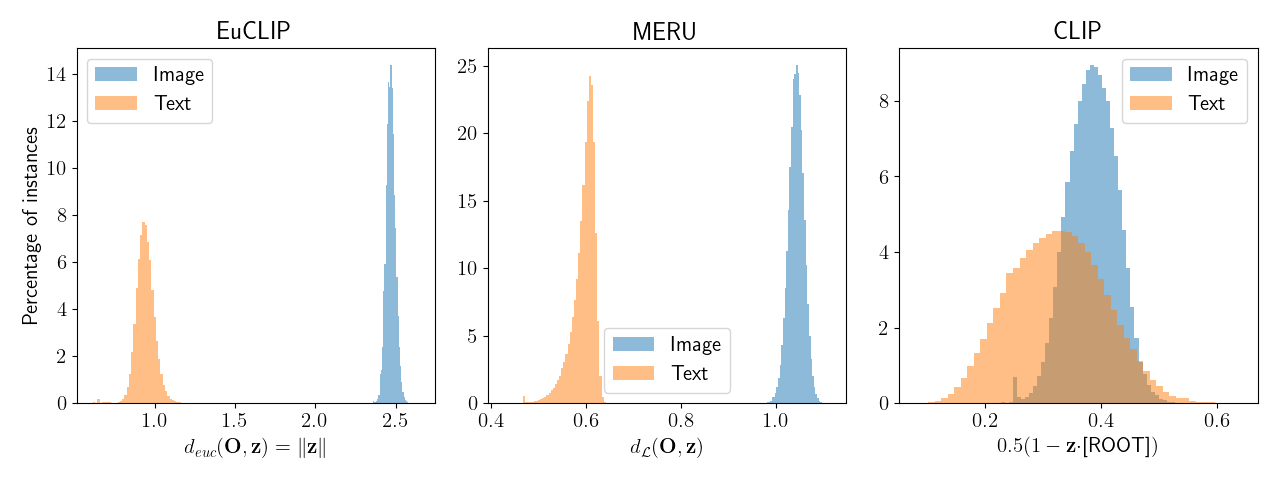}
    \caption{Distribution of embedding distances for ViT-B/16 Models. For EuCLIP and MERU the distances are from the origin $\mathbf{O}$ and for CLIP the distances are from [ROOT], the average of all text and image embeddings after L2 normalization. Note that this scaled ``cosine distance'' $\in [0, 1]$ even though most of the embeddings are no further than $0.5$ from the root, replicating the cone effect\cite{ModalityGap}.}
    \label{fig:VIT-B-16-Norm-Distribution}
\end{figure}

\begin{table}[h!]
\caption{Zero-shot performance for ViT-B/16 models.}
\label{table:1}
\centering
\begin{tabular}{|c|c|c|c|c|c|} 
 \hline
  & & ImageNet &  &  & Average over \\
  & \multirow{-2}{*}{ImageNet} & dist. shifts & \multirow{-2}{*}{VTAB} & \multirow{-2}{*}{Retrieval} & 38 datasets \\
 \hline\hline
 EuCLIP & \textbf{35.17} & \textbf{27.7}& \textbf{37}& \textbf{26.3}& \textbf{35.8}\\ 
 \hline
 CLIP & 34.73 & 27.2& 35.7& 25.7& 34.9\\
 \hline
 MERU & 33.84 & 26.2& 35.6& 25.6& 34.2\\
 \hline
\end{tabular}
\end{table}
As we can see in Table \ref{table:1}, EuCLIP beats both CLIP and MERU on all zero-shot metrics used for DataComp evaluation. Given the similar scale, perhaps it's worth comparing the results with the ones reported by Desai \etal~\cite{DBLP:conf/icml/DesaiNR0V23}. Their models were trained on the full 12M text-image pairs of RedCaps v1.0 for batch size 2048 $\times$ 120K steps $\approx$ 245.8M total samples seen, whereas our models are trained on 12.3M text-image pairs for the worth of 128M total samples seen. We therefore find our ViT-B/16 models' performance on ImageNet (CLIP 34.73\%, MERU 33.84\%) consistent with that of theirs (CLIP 37.9\%, MERU 37.5\%). Qualitatively, we can see the expected embedding space structures in Figure \ref{fig:VIT-B-16-Norm-Distribution} by plotting the distances of all training data embeddings from [ROOT], defined\cite{DBLP:conf/icml/DesaiNR0V23} as the origin $\mathbf{O}$ for EuCLIP and MERU and the average of all text and image embeddings for CLIP. The text embeddings are driven towards the origin $\mathbf{O}$ and the image embeddings are driven away from the origin $\mathbf{O}$ by the entailment loss for both EuCLIP and MERU, while they remain overlapped for CLIP.

\begin{figure}[h!]
\minipage{0.24\textwidth}
  \includegraphics[width=\linewidth]{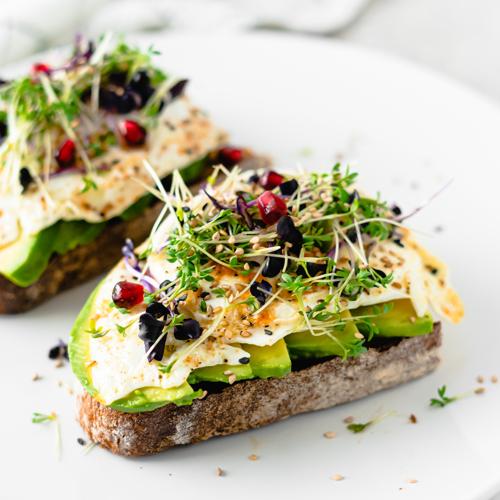}
\endminipage\hfill
\minipage{0.24\textwidth}
  \includegraphics[width=\linewidth]{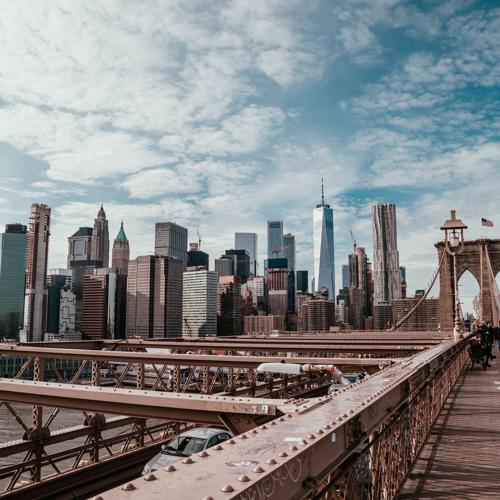}
\endminipage\hfill
\minipage{0.24\textwidth}
  \includegraphics[width=\linewidth]{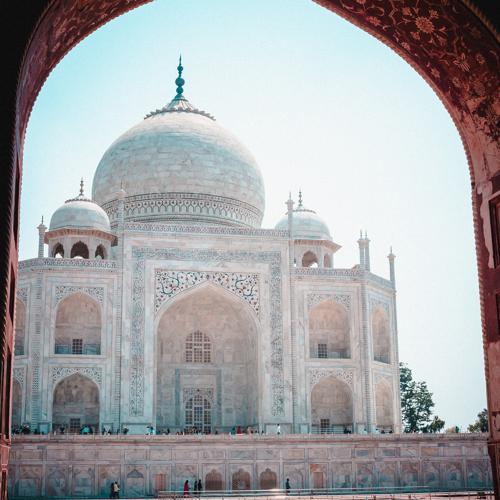}
\endminipage
\minipage{0.24\textwidth}
  \includegraphics[width=\linewidth]{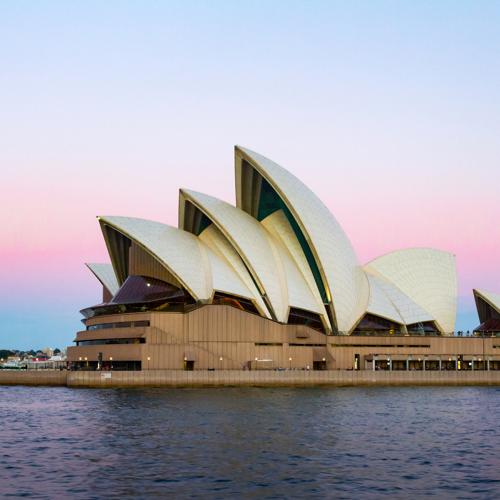}
\endminipage
\caption{Example images from the MERU repository.}\label{fig:image_assets}
\end{figure}

We also perform the image traversals described by Desai \etal~\cite{DBLP:conf/icml/DesaiNR0V23} using the same image (Figure \ref{fig:image_assets}) and text assets. In these image traversals, we linearly interpolate between the image embeddings and [ROOT] and retrieve the closest caption to the interpolated points. We find that none of the 3 models retrieves significantly more distinct captions along such path (Table \ref{table:all_image_traversal}). The result for MERU remains unchanged even if we filter for captions that entail the interpolated points regardless of the minimum radius $K \in [0.1, 0.8]$ used for filtering. Further representation hierarchy only emerges with EuCLIP and adjusted value of $K$, \eg $K = 0.8$ (Table \ref{table:euclip_image_traversal}). For more image traversal details and results, see Supplementary Material \ref{sec:image_traversals}.

\begin{table}[h!]
\caption{Distinct captions retrieved along the paths of image traversal with EuCLIP, CLIP, and MERU for images in Figure \ref{fig:image_assets}. Captions near the top are closer to the image embeddings while captions near the bottom are closer to [ROOT]. From top to bottom, the captions become more and more generic and always end with [ROOT] itself as expected.}
\label{table:all_image_traversal}
\centering
\begin{tabular}{|c|c|c||c|c|c|} 
 \hline
 EuCLIP & CLIP & MERU & EuCLIP & CLIP & MERU \\
 \hline\hline
 \multicolumn{2}{|c|}{avocado toast served} & avocado & brooklyn & photo of brooklyn & brooklyn \\ 
 \multicolumn{2}{|c|}{on white plate} & toast & bridge & bridge, new york & bridge\\
 \hline
 healthy eating & food photography & healthy eating & cityscape & urban & skyline \\
 \hline
 food & $\downarrow$ & $\downarrow$ & $\downarrow$ & $\downarrow$ & $\downarrow$ \\
 \hline
 \multicolumn{3}{|c||}{[ROOT]} & \multicolumn{3}{c|}{[ROOT]} \\
 \hline
 \hline
 EuCLIP & CLIP & MERU & EuCLIP & CLIP & MERU \\
 \hline\hline
 islamic & taj & taj mahal & \multicolumn{3}{c|}{sydney} \\ 
 architecture & mahal & through an arch & \multicolumn{3}{c|}{opera house} \\
 \hline
 tourist spot & landmark & $\downarrow$ & sydney & $\downarrow$ & australia \\
 \hline
 $\downarrow$ & $\downarrow$ & $\downarrow$ & $\downarrow$ & $\downarrow$ & $\downarrow$ \\
 \hline
 \multicolumn{3}{|c||}{[ROOT]} & \multicolumn{3}{c|}{[ROOT]} \\
 \hline
\end{tabular}
\end{table}

\begin{table}[h!]
\caption{Distinct captions retrieved along the paths of image traversal with EuCLIP but with $K = 0.8$, for the same images in Figure \ref{fig:image_assets}. More distinct captions are retrieved than any of the 3 models in Table \ref{table:all_image_traversal}, revealing more hierarchical structure of the embedding space.}
\label{table:euclip_image_traversal}
\centering
\begin{tabular}{|c||c||c||c|}
 \hline
 healthy eating & brooklyn bridge & islamic architecture & sydney \\
 \hline
 food & skyline & tourist attraction & scenery \\
 \hline
 blooming flowers & cityscape & tourist spot & cityscape \\
 \hline
 kitchen & $\downarrow$ & town & $\downarrow$ \\
 \hline
 $\downarrow$ & $\downarrow$ & cityscape & $\downarrow$ \\
 \hline
 \multicolumn{4}{|c|}{[ROOT]} \\
 \hline
\end{tabular}
\end{table}

\subsection{ViT-B/32 Model Experiments}

\begin{figure}
    \centering
    \includegraphics[width=0.75\linewidth]{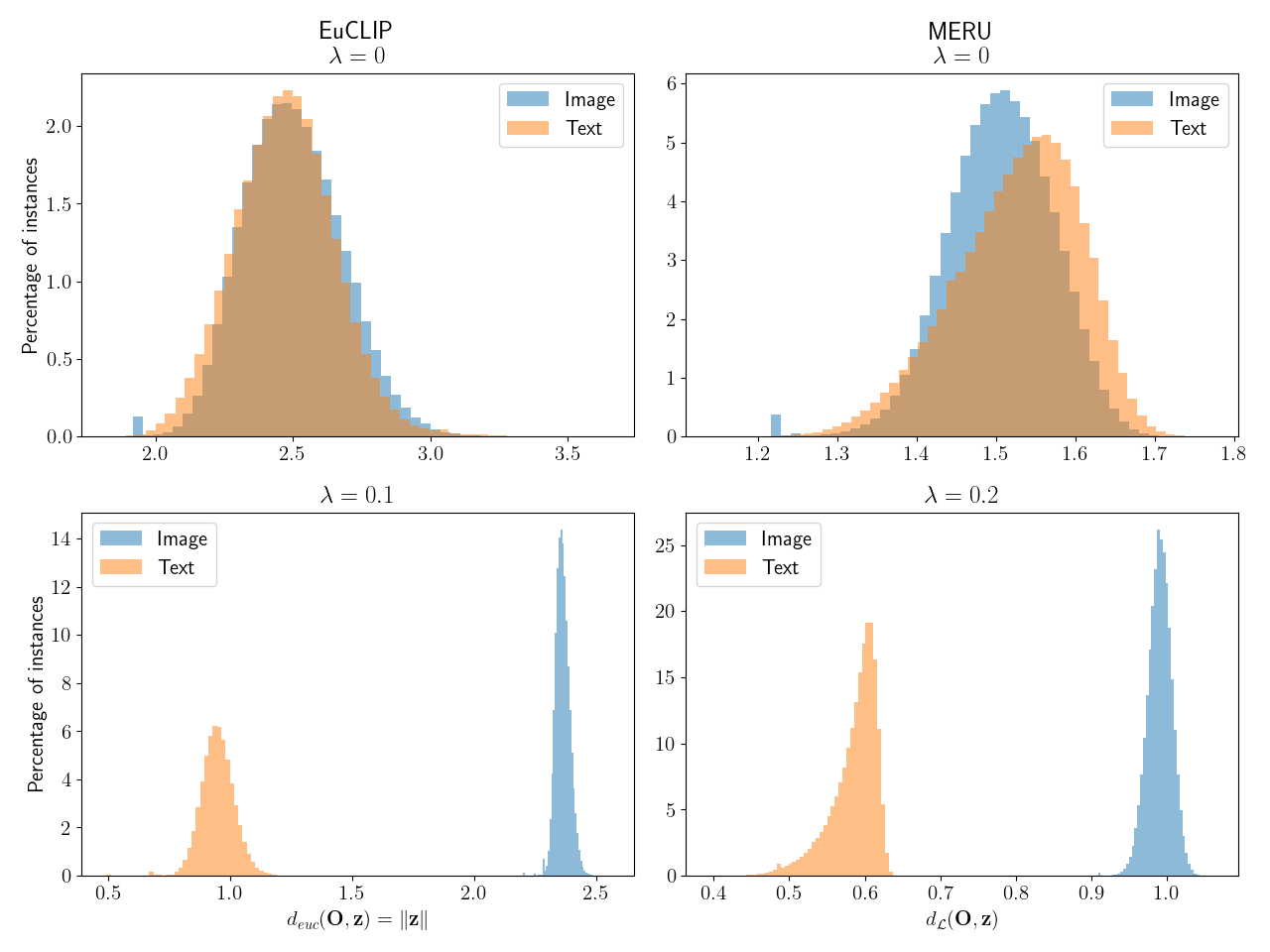}
    \caption{Distribution of embedding distances from the origin $\mathbf{O}$ for ViT-B/32 Models, EuCLIP (left) vs. MERU (right) and with $\lambda = 0$ (upper) vs. $\lambda > 0$ (lower). As the upper panels show, text and image embeddings do not spontaneously separate. Such ``modality gap''~\cite{Ramasinghe2024} only emerges with entailment loss.}
    \label{fig:ViT-B-32-Norm-Distribution}
\end{figure}

Table \ref{table:ViT-B-32-Models} represents the full set of relevant ViT-B/32 models we train at the DataComp medium scale. With the possible exception of the effect of entailment loss on retrieval tasks, we can see that Euclidean geometry, distance squared $d^2$ logit, no final LN (no-ln), and training with entailment loss hold advantage over hyperbolic geometry, distance $d$ logit, unmodified encoders, and training without entailment loss respectively. MERU is a rather unfavorable combination and with no final LN ($d$, no-ln, $\lambda \in [0, 0.2]$) turns out to be unstable while EuCLIP without entailment loss ($d^2$, no-ln, $\lambda$=0) already matches the performance of CLIP. We observe that in the $n = 512$ dimensional embedding space here, the volume of the $\epsilon$-ball around an embedding grows $\sim O(\epsilon^{512})$, so the exponential volume growth of the $\epsilon$-ball in hyperbolic geometry offers little advantage in practice. The observation of the advantage of hyperbolic embedding space over Euclidean embedding space when $n$ is small and conversely its diminishing when $n$ is large has been made in several studies\cite{DBLP:journals/corr/abs-2109-07488,DBLP:conf/cvpr/KhrulkovMUOL20}. With $n$ often in the hundreds, we expect the latter case to become more common. We further observe that throughout our training process the curvature parameter $c$ consistently decreases and all our hyperbolic geometry models at the medium scale and all the MERU checkpoints published by Desai \etal~\cite{DBLP:conf/icml/DesaiNR0V23} end up with the clamped minimum $c = 0.1$, replicating the finding of\cite{Ramasinghe2024} and demonstrating the unfavorability of hyperbolicity. While we have less insight on the role of entailment loss with respect to the model's performance, we can answer the long-standing question on whether the separation between text and image embeddings emerges spontaneously~\cite{spontaneous_embedding_separation_q} in the negative by comparing the embedding distance distribution of EuCLIP and MERU to that of their counterparts trained without entailment loss. As we can see in Figure \ref{fig:ViT-B-32-Norm-Distribution}, the distance distributions for text and image embeddings remain overlapped and in fact appear identical for EuCLIP with $\lambda = 0$. Interestingly, the observations on the model performance do not hold for the small scale (Supplementary Material \ref{sec:Small-ViT-B-32-Models}) where CLIP and elliptic geometries have the advantage. We hypothesize that with such limited training data, more restricted $(n-1)$-sphere $S^{n - 1}$ embedding space forces the model to learn the latent structure instead of memorization, and the 3125-step training budget prevents grokking~\cite{DBLP:journals/corr/abs-2201-02177} for Euclidean and hyperbolic geometries.

We answer the questions regarding the final LN in the next section.

\begin{table}[h!]
\caption{Zero-shot performance for medium scale ViT-B/32 models.}
\label{table:ViT-B-32-Models}
\centering
\begin{tabular}{|c|c|c|c|c|c|c|} 
 \hline
  & & & ImageNet &  &  & Average over \\
  \multirow{-2}{*}{Geometry} & \multirow{-2}{*}{Variant} & \multirow{-2}{*}{ImageNet} & dist. shifts & \multirow{-2}{*}{VTAB} & \multirow{-2}{*}{Retrieval} & 38 datasets \\
 \hline\hline
 \multicolumn{2}{|c|}{CLIP} & 27.92 & 22.1& 32.3& 21.1& 31.1\\
 \hline
 \multicolumn{2}{|c|}{Elliptic} & 27.80 & 22.3& \textbf{33.5} & 21.3 & \textbf{32.0} \\
 \hline
  & $d^2$, no-ln, $\lambda$=0.2 & 28.41 & \textbf{23.0} & 32.6& 20.9& 31.2\\
 \cline{2-7}
  & EuCLIP  & \textbf{28.97} & \textbf{23.0} & \textbf{33.5} & 21.0 & \underline{31.8}\\
 \cline{2-7}
  & $d^2$, no-ln, $\lambda$=0 & 27.66 & 22.1& 33.0& \underline{21.6} & 31.1\\
 \cline{2-7}
  & $d$, no-ln, $\lambda$=0 & 26.03 & 20.4& 31.2& 20.5& 29.8\\
 \cline{2-7}
 \multirow{-5}{*}{Euclidean} & $d$, ln, $\lambda$=0 & 25.09 & 20.2& 32.4& \underline{21.6} & 30.9\\
 \hline
  & $d^2$, no-ln, $\lambda$=0.2 & 27.51 & 22.1& 32.4& 20.8& 30.9\\
 \cline{2-7}
  & $d^2$, no-ln, $\lambda$=0 & 25.71 & 20.1& 32.1& 21.1& 30.3\\
 \cline{2-7}
  & MERU & 26.88 & 21.6& \underline{33.4} & 20.8& 30.8\\
 \cline{2-7}
 \multirow{-4}{*}{Hyperbolic} & $d$, ln, $\lambda$=0 & 24.58 & 19.5& 31.0& \textbf{21.7} & 29.6\\
 \hline
\end{tabular}
\end{table}

\subsection{Final LN Ablation}
The experiments in the previous section are controlled in the sense that only one change is made at a time. This however leaves the question of the effect size of final LN removal and whether it's additive open. We therefore run two more ablation experiments by restoring the final LN of the EuCLIP models. As we can see in Table \ref{table:4}, restoring the final LN alone drastically impacts the model performance. In fact, ViT-B/16 final-ln barely outperforms ViT-B/32 no-ln with 4 times the number of patches. If we inspect the model architecture of the text and image encoders, we can see that they use the version of LN with learnable per-element affine parameters, weight $\mathbf{w}$ and bias $\mathbf{b}$, followed by a linear layer parameterized as projection matrix $\mathbf{P}$. The final embedding generated by the encoders therefore can be written as $\mathbf{P}(\mathrm{diag}(\mathbf{w}) \mathbf{x} + \mathbf{b})$ where $\mathbf{x}$ is a normalized vector with mean 0 and variance 1, with implied L2 norm $\|\mathbf{x}\| = \sqrt{n}$. By linearity, the final embedding can be written as $\mathbf{W}\mathbf{x} + \mathbf{b}'$ where $\mathbf{W} = \mathbf{P}\mathrm{diag}(\mathbf{w})$ and $\mathbf{b}' = \mathbf{P}\mathbf{b}$. Furthermore, all matrices are almost diagonalizable over $\mathbb{C}$\cite{3146457}. The implication for real matrix $\mathbf{W}$ is that almost all of them can be put in real Jordan form consists of only $2 \times 2$ rotation-scaling Jordan blocks or diagonal values, which in turn imply that the linear transformation it represents can be described by orthonormal basis $e_1, e_2, \dots, e_n$, complex eigenvalues $\lambda_1, \lambda_2, \dots, \lambda_n$, and the rotated orthonormal basis $e'_1, e'_2, \dots, e'_n$. So the final embedding effectively must reside on the hyperellipsoid spanned by $\sqrt{n}|\lambda_i|e'_i$, shifted by $\mathbf{b}'$. The degree of freedom lost due to the final LN cannot be recovered.

The closest parallel we can draw is from \cite{DBLP:journals/corr/abs-2109-07488}, in which Bansal-Benton pointed out that Euclidean embeddings become competitive when L2 norm clipping is removed or relaxed to a larger maximum. However, our finding may have wider implications. This earlier form of Pre-LN Transformer, Pre-LN Transformer without the final LN, is the only transformer architecture whose final output doesn't go through LN. Unless further non-linearity is applied to their output, all the other transformer architectures will suffer the same loss of degree of freedom if the norm of the output is relevant. People therefore may have been rejecting novel model architectures or loss terms unfairly due to poor performance resulting from such incompatibility.

\begin{table}[h!]
\caption{Zero-shot performance for EuCLIP final LN ablation.}
\label{table:4}
\centering
\begin{tabular}{|c|c|c|c|c|c|c|} 
 \hline
  & & & ImageNet &  &  & Average over \\
  \multirow{-2}{*}{Model} & \multirow{-2}{*}{Variant} & \multirow{-2}{*}{ImageNet} & dist. shifts & \multirow{-2}{*}{VTAB} & \multirow{-2}{*}{Retrieval} & 38 datasets \\
 \hline\hline
  & no-ln & \textbf{35.17} & \textbf{27.7}& \textbf{37}& \textbf{26.3}& \textbf{35.8}\\
 \cline{2-7}
 \multirow{-2}{*}{ViT-B/16} & ln & 29.48 & 21.5& 31.9& 21.5& 31.0\\
 \hline
  & no-ln & 28.97 & 23.0& 33.5& 21.0& 31.8\\
 \cline{2-7}
 \multirow{-2}{*}{ViT-B/32} & ln & 22.22 & 16.5& 28.8& 18.0& 27.2\\
 \hline
 \end{tabular}
\end{table}

\section{Conclusion}
We have systematically tested alternative embedding geometries and softmax logits for contrastive language-image pre-training, with emphasis on the unexplored but intuitive Euclidean geometry. We find that the final LN of most transformer architectures results in loss of degree of freedom and severely impacts model performance when the norm of the output carries information. We find that the combination of Euclidean geometry, distance squared $d^2$ logit, no final LN, and training with Euclidean entailment loss (EuCLIP) results in models that match or outperform CLIP, add no additional trainable parameters, and support hierarchical relationships at least as well as more complicated MERU. Furthermore, Euclidean distance is better supported by nearest-neighbor libraries like the FAISS library~\cite{DBLP:journals/corr/abs-2401-08281} than its hyperbolic counterpart even disregarding the latter's parameterization by curvature parameter $c$. We therefore believe EuCLIP should be considered for further scaling up and applications.

\subsection{Limitations}
Due to copyright, text-image pair datasets usually only contain links to the images instead of the images themselves and DataComp is no exception. The images may be taken down or of restricted access to begin with, resulting in dead links that preclude full reproducibility~\cite{DBLP:journals/corr/abs-2310-03193}. Indeed, we have only been able to download $<90$\% of the images at their respective DataComp scales and our CLIP models get close but do not match the reference performance. Fully public and sizable datasets or a centralized setting in which the data is permanent and researchers submit code and training hyperparameters can alleviate this issue.

We do not fully understand the entailment loss, its interactions with the InfoNCE loss, or why it improves zero-shot classification but not retrieval. We are also surprised by the fact that entailment loss does not result in ``\textit{concentric, high-dimensional rings} around [ROOT]''\cite{DBLP:conf/icml/DesaiNR0V23} and therefore call the previous definition of [ROOT] into question. In fact, the average embedding deviates further from the origin $\mathbf{O}$ in models trained with entailment loss than without (Supplementary Material \ref{sec:embedding_dist_dist}). It is entirely possible that there is a better alternative to train the model to have hierarchical representations.

Finally, since we adopt the zero-shot evaluation protocol of DataComp, we left the performance of EuCLIP for linear probe, fine-tuning, or downstream application unexamined. The DataComp study justifies this evaluation protocol with a strong rank correlation between zero-shot and linear probe performance~\cite{DBLP:conf/nips/GadreIFHSNMWGZO23}, but it is less clear whether such rank correlation continues to hold for models with different underlying geometries.

\bibliographystyle{splncs04}
\bibliography{main}

\newpage
\appendix
\addcontentsline{toc}{section}{Appendices}
\section*{Supplementary Material}
\section{Source Code}
\label{sec:source_code}
All the models presented in the main paper can be trained and evaluated with the following repositories:

\begin{enumerate}
\item Modified \href{https://github.com/EIFY/open_clip/tree/euclip/}{open\_clip}
\item Modified \href{https://github.com/EIFY/datacomp/tree/euclip/}{datacomp}
\item Modified \href{https://github.com/EIFY/CLIP_benchmark/tree/euclip/}{CLIP\_benchmark}, as dependency of 2
\end{enumerate}
In 1, \texttt{class CLIP} in \href{https://github.com/EIFY/open_clip/tree/euclip/src/open_clip/model.py}{model.py} implements EuCLIP / CLIP / MERU, depending on the \texttt{geometry}. \texttt{METRICS} in \href{https://github.com/EIFY/open_clip/tree/euclip/src/open_clip/loss.py}{loss.py} in turn defines all of the distance metrics tested. Evaluation functions of 2 and 3 then evoke \texttt{METRICS}, \eg \href{https://github.com/EIFY/datacomp/tree/euclip/eval_utils/wino_eval.py}{wino\_eval.py} of 2 and  \href{https://github.com/EIFY/CLIP_benchmark/tree/euclip/clip_benchmark/metrics/zeroshot_classification.py}{zeroshot\_classification.py} of 3. 

We then create dedicated branch of open\_clip to compute the embedding average and distance distribution using modified \href{https://github.com/nahidalam/open_clip/tree/euclip/src/training/train.py}{\texttt{evaluate()}}. Image traversal is then performed with \href{https://github.com/nahidalam/datacomp/tree/traversal/image_traversal.py}{image\_traversal.py}, using the calculated embedding average [ROOT] when necessary. Finally, we modify \href{https://github.com/EIFY/meru/tree/no_filtering/scripts/image_traversals.py}{image\_traversals.py} from the original \href{https://github.com/facebookresearch/meru}{meru} repository to perform image traversals using their published model checkpoints to compare the results.

\section{Image Traversals: More Details and Results}
\label{sec:image_traversals}
For image traversal, we follow the practice of~\cite{DBLP:conf/icml/DesaiNR0V23}.
\subsection{Method}
We calculate the image embedding $\mathbf{y}$ and linearly interpolate between $\mathbf{y}$ and the root\footnote{The root is the origin $\mathbf{O}$ for EuCLIP and MERU, and the embedding average [ROOT] of all the training text and images after L2 normalization for CLIP.} in 50 equally spaced steps, inclusive of $\mathbf{y}$ and the root themselves:

\begin{itemize}
\item For CLIP, the image embedding is L2-normalized before linear interpolation, and the resulting interpolated steps are L2-normalized again.
\item For EuCLIP, interpolated steps are used as they are.
\item For MERU, interpolation is done before exponential lifting and the resulting interpolated steps are then exponentially lifted.

\end{itemize}

Optionally, for EuCLIP and MERU, we first filter for the captions whose embedding $\mathbf{x}$ entails the interpolated step embedding, \ie the captions whose $\mathbf{x}$ satisfies $\mathcal{L}_{entail}(\mathbf{x}, \mathbf{y}_{step,i}) = 0$ before retrieving for the nearest-neighbor caption, $\argmax_{\mathbf{x}} \mathrm{sim}(\mathbf{x}, \mathbf{y}_{step,i})$ in their respective geometry. We use different values of the minimum radius $K$ to calculate the entailment loss $\mathcal{L}_{entail}(\mathbf{x}, \mathbf{y}_{step,i})$, in addition to the value used for training. Segments of the interpolation often retrieve the same captions, so we filter out the duplicated captions and only count deduplicated captions for our result.

\subsection{Data}
For maximum reproducibility, we use the same 60 randomly selected images collected from \href{https://www.pexels.com/}{pexels.com}. Some of the links are now dead, but we find that the TeX source of~\cite{DBLP:conf/icml/DesaiNR0V23} retains $500 \times 500$ thumbnails that are of sufficient resolution for our ViT encoders. For candidate captions, we also reuse \href{https://github.com/facebookresearch/meru/blob/main/assets/pexels_text.json}{pexels\_text.json} of the MERU repo. We then perform the same prompt formatting by keeping the original captions, formatting noun tags as \texttt{`a photo of \{\}.'}, and formatting adjective tags as \texttt{`this photo is \{\}.'}.

\subsection{Result}

\begin{figure}
    \centering
    \includegraphics[width=1.0\linewidth]{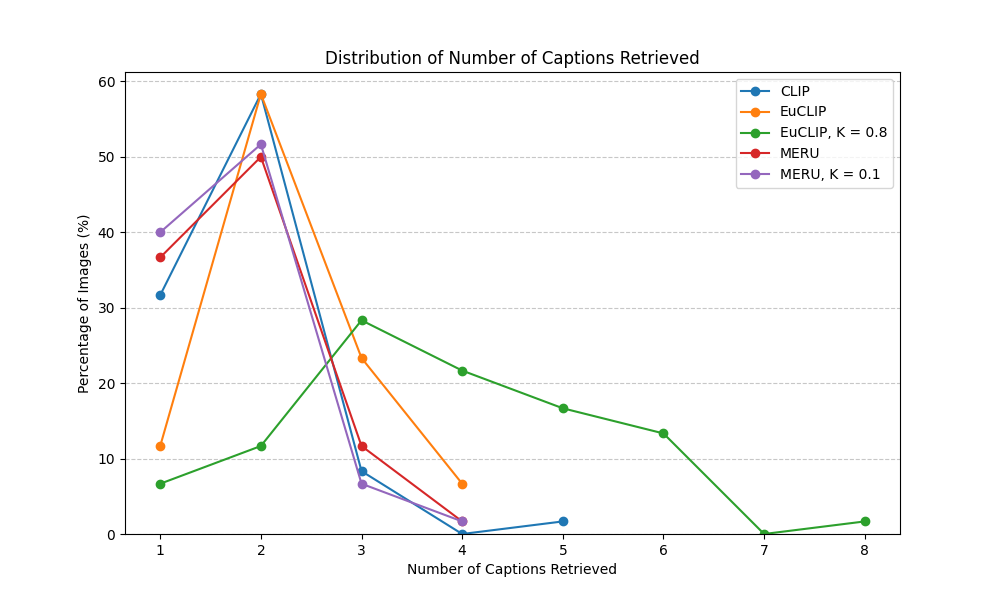}
    \caption{Distribution of number of captions retrieved.}
    \label{fig:unique_counts_distribution}
\end{figure}

Here are the averages (Table \ref{table:avg_captions_retrieved}) and distributions (Figure \ref{fig:unique_counts_distribution}) of the number of captions retrieved per image for our ViT-B/16 models from Table \ref{table:1} with no entailment filtering or various values of $K$ of interest, excluding the root itself.

We can see that EuCLIP with entailment filtering using $K = 0.8$ retrieves the most captions along the interpolations in average, followed by EuCLIP without entailment filtering. Interestingly, running EuCLIP with entailment filtering using lower minimum radius, $K \in [0.3, 0.7]$, results in zero captions retrieved other than the root, possibly because of the entailment loss weight $\lambda = 0.1$ used for training and how its entailment loss $\mathcal{L}_{entail} \neq 0$ with minimum radius $K = 0.3$ during training. In contrast, the result of image traversal barely changes for MERU, with or without entailment filtering, using all the values of $K \in [0.1, 0.8]$ tested. Since $K = 0.1$ is its value in training, it is the lowest sensible value to use and results in the strongest filtering, we consider it representative. The fact that even $K = 0.1$ entailment filtering barely changes the image traversal result for MERU suggests that hyperbolic entailment loss hasn't been very effective in helping the model learn hierarchical representation. For side-by-side comparison, we then test the MERU model from~\cite{DBLP:conf/icml/DesaiNR0V23} at the same model scale, \texttt{MERU ViT-B/16}, using the published checkpoint (Table \ref{table:meru_vit_b_16_avg_captions_retrieved}). It retrieves significantly more captions than our comparable MERU model, but entailment filtering still only results in limited increase.

\begin{table}[h!]
\caption{Average number of captions retrieved per image.}
\label{table:avg_captions_retrieved}
\centering
\begin{tabular}{|c|c|} 
 \hline
  & Average number of captions retrieved \\
 \hline\hline
 CLIP & 1.817 \\
 \hline
 EuCLIP & 2.25 \\
 \hline
 EuCLIP, K = 0.8 & \textbf{3.783} \\
 \hline
 MERU & 1.783 \\
 \hline
 MERU, K = 0.1 & 1.7 \\
 \hline
\end{tabular}
\end{table}

\begin{table}[h!]
\caption{Average number of captions retrieved per image by \texttt{MERU ViT-B/16} from~\cite{DBLP:conf/icml/DesaiNR0V23}. We cannot test \texttt{CLIP ViT-B/16} here because its root is missing from the model checkpoint.}
\label{table:meru_vit_b_16_avg_captions_retrieved}
\centering
\begin{tabular}{|c|c|} 
 \hline
  & Average number of captions retrieved \\
 \hline\hline
 No filtering & 3.383 \\
 \hline
 K = 0.1 & 3.617 \\
 \hline
\end{tabular}
\end{table}

Perhaps it is worth considering why the results from~\cite{DBLP:conf/icml/DesaiNR0V23} don't seem to replicate. We have the following 3 hypotheses, in decreasing order of likelihood:

\begin{enumerate}
    \item  Bias of the dataset: In particular, RedCaps~\cite{desai2021redcaps} is not just a dataset of text-image pairs. It has a subreddit field and in~\cite{DBLP:conf/icml/DesaiNR0V23} the caption is augmented with 0.5 probability to be \texttt{`\{subreddit\}: \{caption\}'} during training for both CLIP and MERU. Such data may be particularly helpful for model with built-in hierarchical representation support.
    \item  Variance of the dataset: That is, if we construct a new version of DataComp datasets with newer Common Crawl or new version of RedCaps with newer reddit images, the results may change again.
    \item  Difference in numerical smoothing implementation, including but not limited to the half-aperture calculation (Section \ref{sec:hyperbolic_entailment_loss}).
\end{enumerate}

Lastly, we would like to emphasize that while number of captions retrieved through image traversal constitutes a metric, it does not address the question of how relevant and how `generic' the retrieved captions are to the image in question, a question that is harder to answer objectively. Therefore, the use of image traversal to assess the model's hierarchical representation remains mostly qualitative. In the spirit of full transparency, here is the image traversal results of the remaining 56 of the 60 examples from~\cite{DBLP:conf/icml/DesaiNR0V23} with the ViT-B/16 EuCLIP model and entailment filtering with $K = 0.8$:

\begin{tabular}{|c||c||c||c|}
 \hline
\includegraphics[width=25mm, height=25mm]{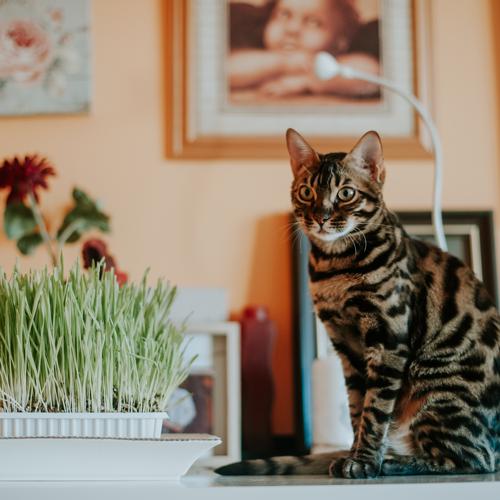} & \includegraphics[width=25mm, height=25mm]{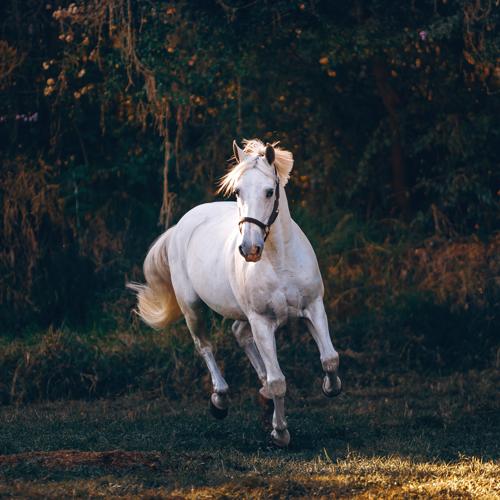} & \includegraphics[width=25mm, height=25mm]{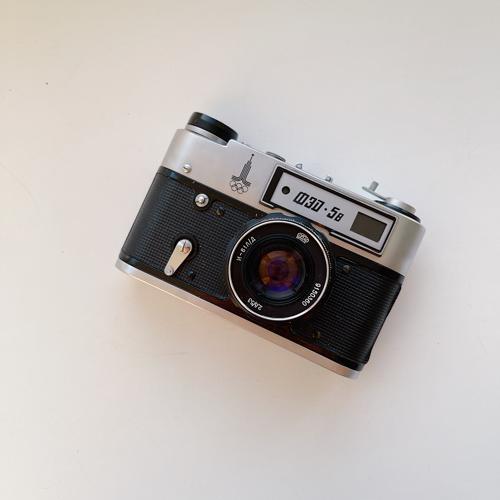} & \includegraphics[width=25mm, height=25mm]{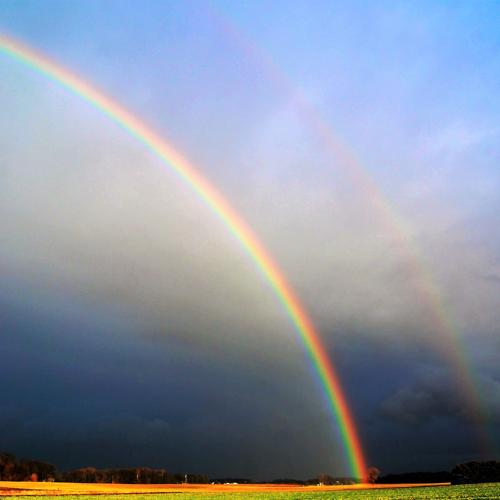} \\
 \hline
cat & horse & photo camera & rainbow \\
 \hline
domestic animal & animal photography & pets & sky \\
 \hline
$\downarrow$ & dog & domestic dog & sky background \\
 \hline
$\downarrow$ & animal & national park & landscape \\
 \hline
$\downarrow$ & domestic dog & $\downarrow$ & scenery \\
 \hline
$\downarrow$ & national park & $\downarrow$ & national park \\
 \hline\multicolumn{4}{|c|}{[ROOT]} \\
 \hline
\end{tabular}

\begin{tabular}{|c||c||c||c|}
 \hline
\includegraphics[width=25mm, height=25mm]{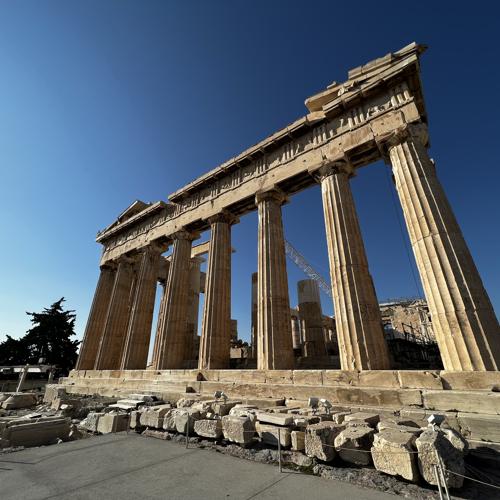} & \includegraphics[width=25mm, height=25mm]{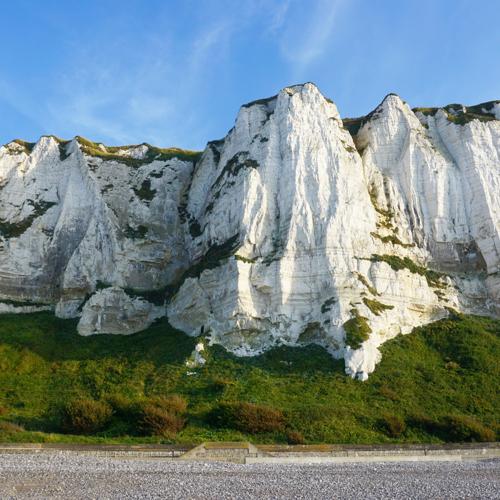} & \includegraphics[width=25mm, height=25mm]{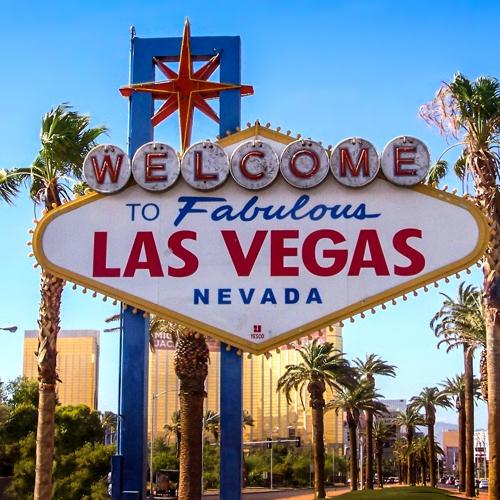} & \includegraphics[width=25mm, height=25mm]{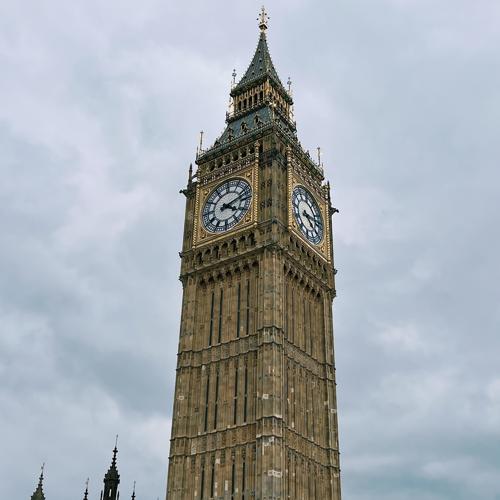} \\
 \hline
athens & cliffs & town & big ben \\
 \hline
\makecell{unesco world \\ heritage site} & geological formation & downtown & \makecell{palace of \\ westminster} \\
 \hline
$\downarrow$ & national park & cityscape & city \\
 \hline
$\downarrow$ & $\downarrow$ & $\downarrow$ & town \\
 \hline\multicolumn{4}{|c|}{[ROOT]} \\
 \hline
\end{tabular}

\begin{tabular}{|c||c||c||c|}
 \hline
\includegraphics[width=25mm, height=25mm]{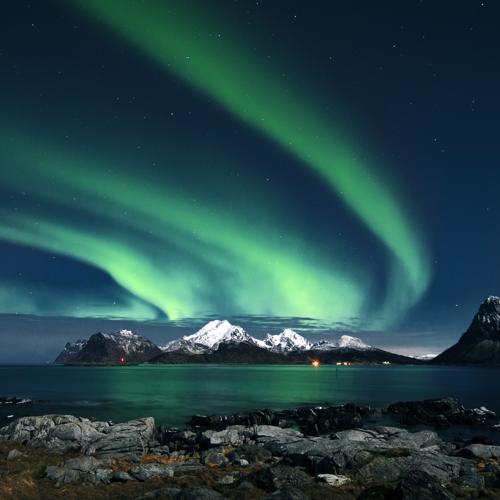} & \includegraphics[width=25mm, height=25mm]{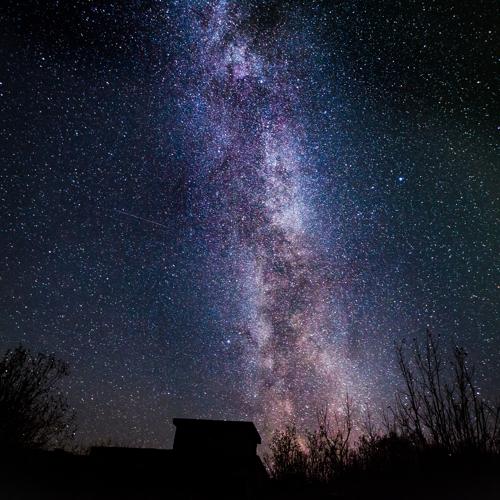} & \includegraphics[width=25mm, height=25mm]{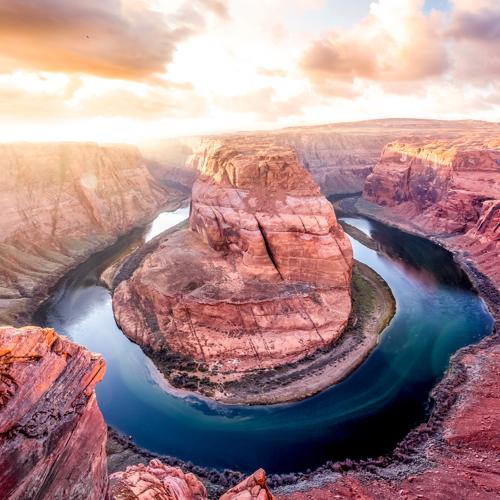} & \includegraphics[width=25mm, height=25mm]{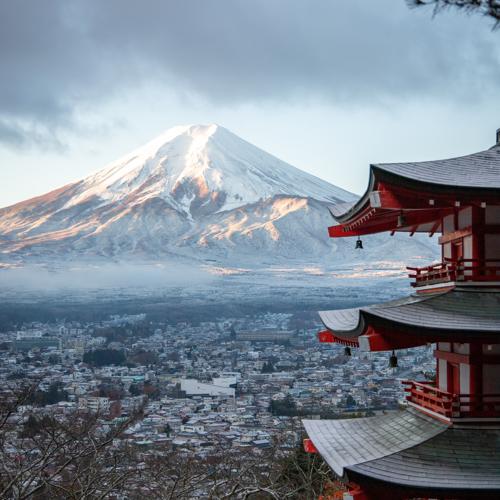} \\
 \hline
northern lights norway & milky way & horseshoe bend & lofoten winter \\
 \hline
northern lights & starry sky & colorado river & scenery \\
 \hline
nature photography & galaxy & scenic & mountains \\
 \hline
landscape & galaxy background & landscape & $\downarrow$ \\
 \hline
mountains & national park & scenery & $\downarrow$ \\
 \hline
$\downarrow$ & $\downarrow$ & national park & $\downarrow$ \\
 \hline\multicolumn{4}{|c|}{[ROOT]} \\
 \hline
\end{tabular}

\begin{tabular}{|c||c||c||c|}
 \hline
\includegraphics[width=25mm, height=25mm]{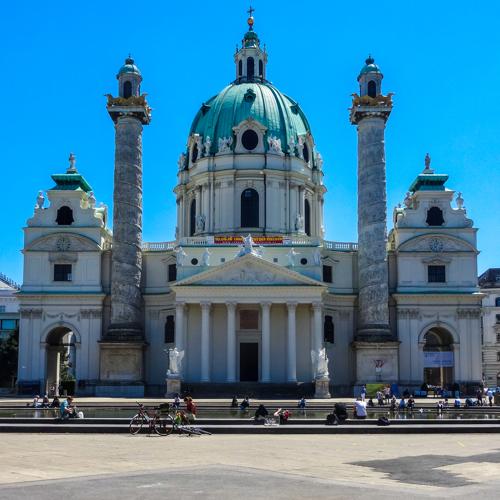} & \includegraphics[width=25mm, height=25mm]{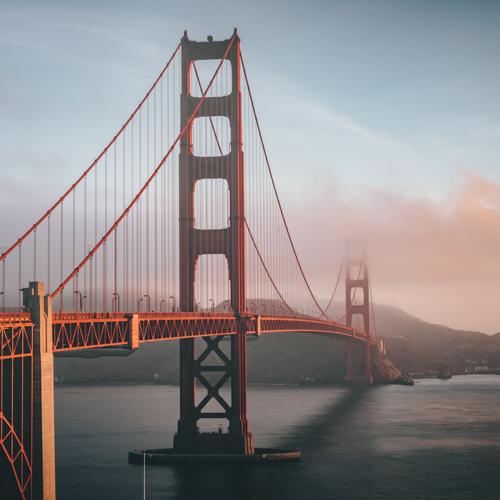} & \includegraphics[width=25mm, height=25mm]{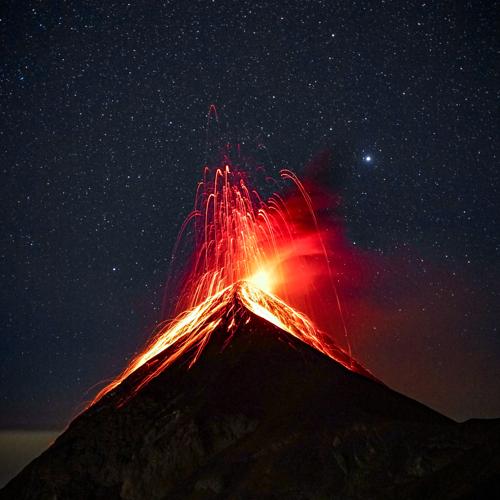} & \includegraphics[width=25mm, height=25mm]{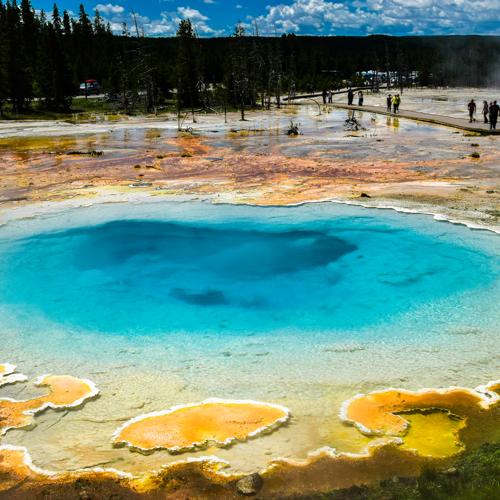} \\
 \hline
famous landmark & golden gate & night sky & colorado river \\
 \hline
karlskirche & san francisco & mountain & idyllic \\
 \hline
tourist attraction & scenery & mountains & destination \\
 \hline
\makecell{unesco world \\ heritage site} & national park & $\downarrow$ & national park \\
 \hline
town & $\downarrow$ & $\downarrow$ & $\downarrow$ \\
 \hline\multicolumn{4}{|c|}{[ROOT]} \\
 \hline
\end{tabular}

\begin{tabular}{|c||c||c||c|}
 \hline
\includegraphics[width=25mm, height=25mm]{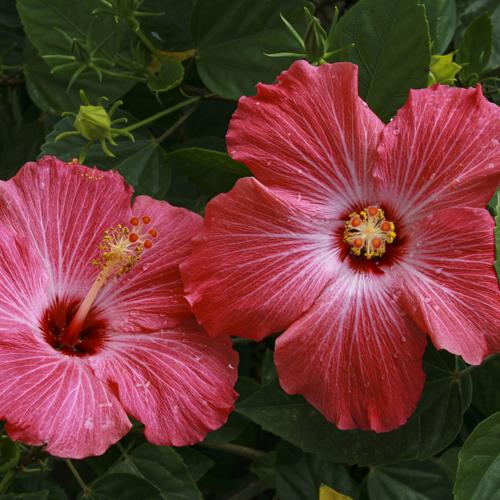} & \includegraphics[width=25mm, height=25mm]{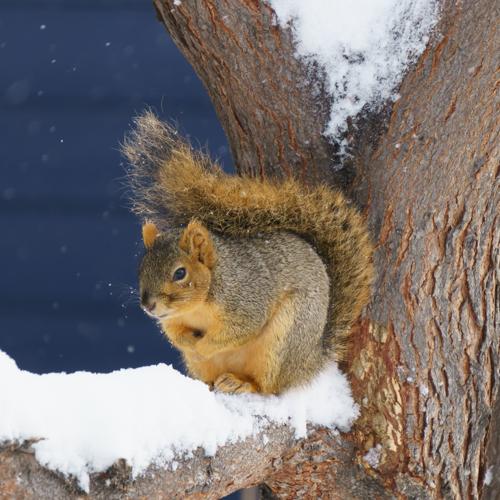} & \includegraphics[width=25mm, height=25mm]{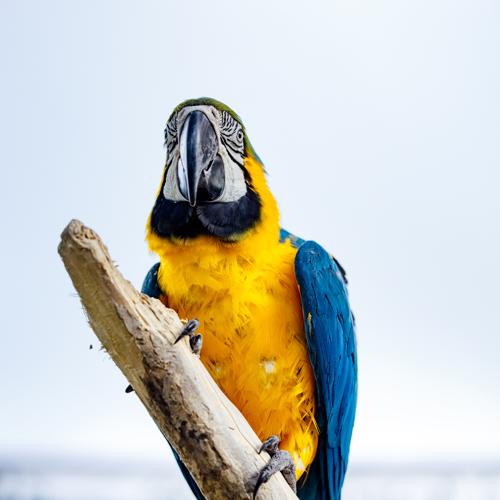} & \includegraphics[width=25mm, height=25mm]{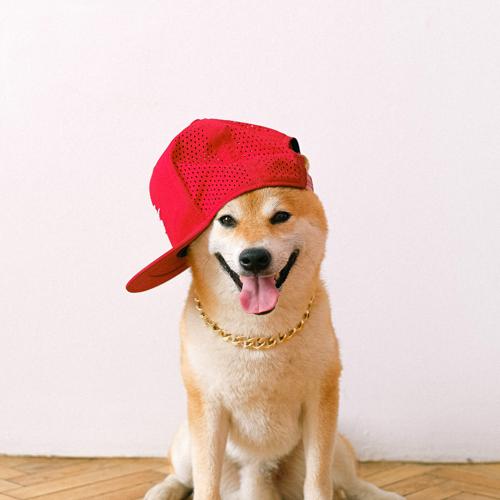} \\
 \hline
red hibiscus in bloom & squirrel & beak & dog \\
 \hline
flower photography & wildlife photography & national park & pets \\
 \hline
blooming flowers & domestic animals & $\downarrow$ & domestic dog \\
 \hline\multicolumn{4}{|c|}{[ROOT]} \\
 \hline
\end{tabular}

\begin{tabular}{|c||c||c||c|}
 \hline
\includegraphics[width=25mm, height=25mm]{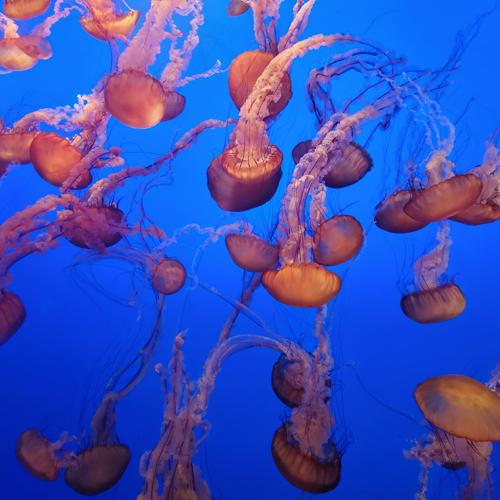} & \includegraphics[width=25mm, height=25mm]{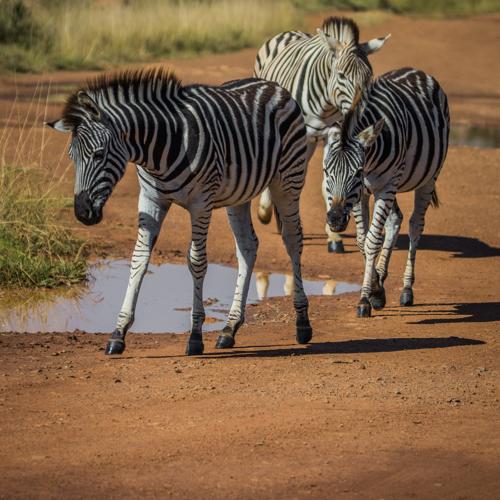} & \includegraphics[width=25mm, height=25mm]{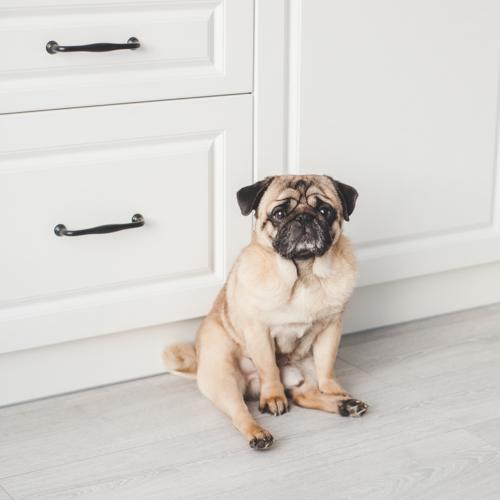} & \includegraphics[width=25mm, height=25mm]{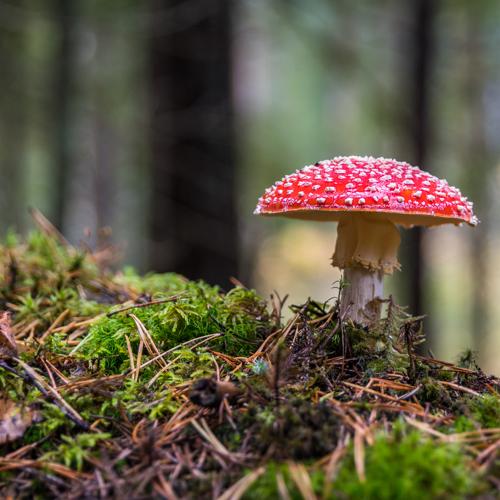} \\
 \hline
sea life & zebras & domestic dog & toadstool \\
 \hline
marine life & galloping & $\downarrow$ & fungus \\
 \hline
reef & wild animals & $\downarrow$ & nature photography \\
 \hline
domestic animals & wildlife photography & $\downarrow$ & spring \\
 \hline
national park & animal & $\downarrow$ & scenery \\
 \hline
$\downarrow$ & national park & $\downarrow$ & blooming flowers \\
 \hline
$\downarrow$ & $\downarrow$ & $\downarrow$ & domestic animals \\
 \hline
$\downarrow$ & $\downarrow$ & $\downarrow$ & mountains \\
 \hline\multicolumn{4}{|c|}{[ROOT]} \\
 \hline
\end{tabular}

\begin{tabular}{|c||c||c||c|}
 \hline
\includegraphics[width=25mm, height=25mm]{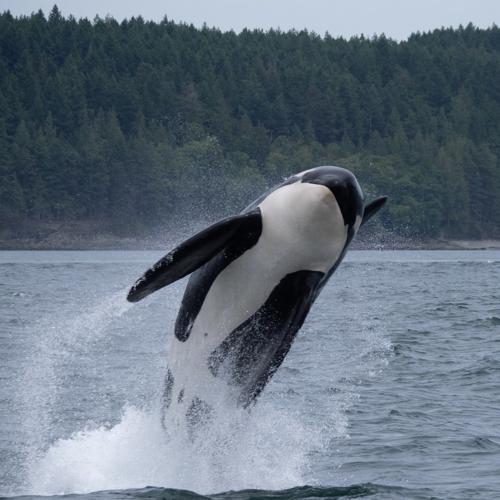} & \includegraphics[width=25mm, height=25mm]{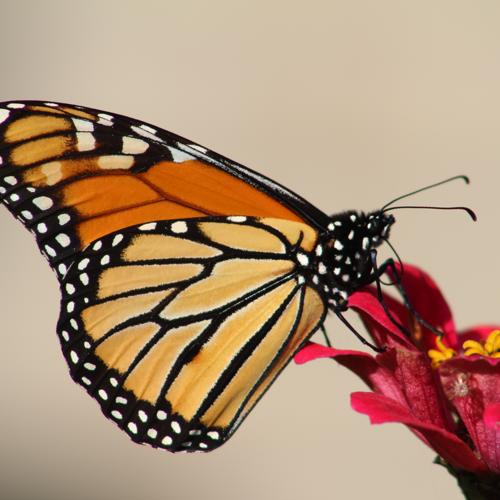} & \includegraphics[width=25mm, height=25mm]{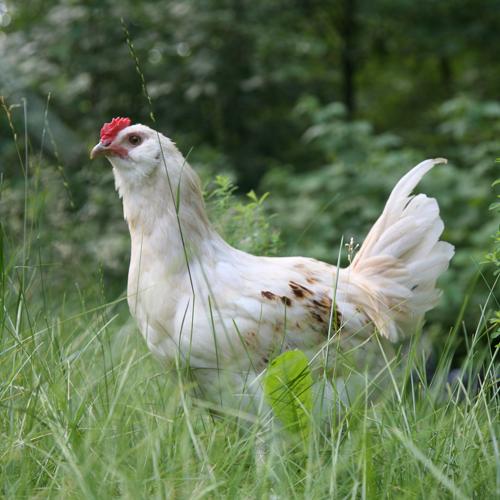} & \includegraphics[width=25mm, height=25mm]{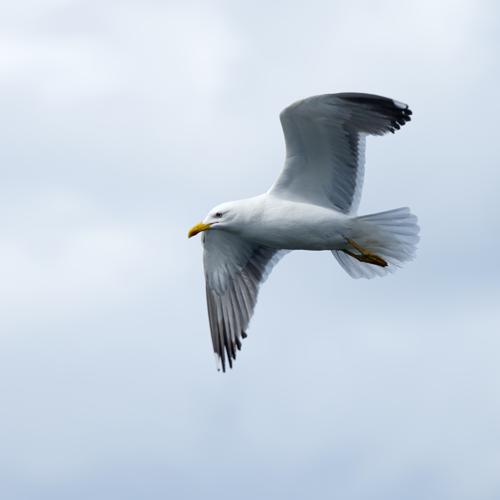} \\
 \hline
whale & butterfly wallpaper & cock & seagull \\
 \hline
beak & butterfly & tranquil & beak \\
 \hline
coast & nature photography & female & national park \\
 \hline
animal & flower photography & wildlife photography & $\downarrow$ \\
 \hline
national park & blooming flowers & domestic animals & $\downarrow$ \\
 \hline
$\downarrow$ & national park & national park & $\downarrow$ \\
 \hline\multicolumn{4}{|c|}{[ROOT]} \\
 \hline
\end{tabular}

\begin{tabular}{|c||c||c||c|}
 \hline
\includegraphics[width=25mm, height=25mm]{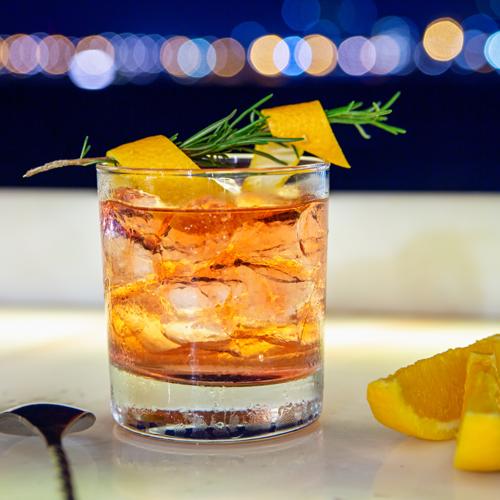} & \includegraphics[width=25mm, height=25mm]{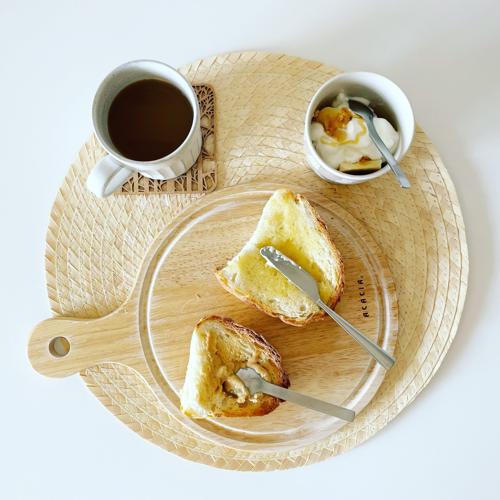} & \includegraphics[width=25mm, height=25mm]{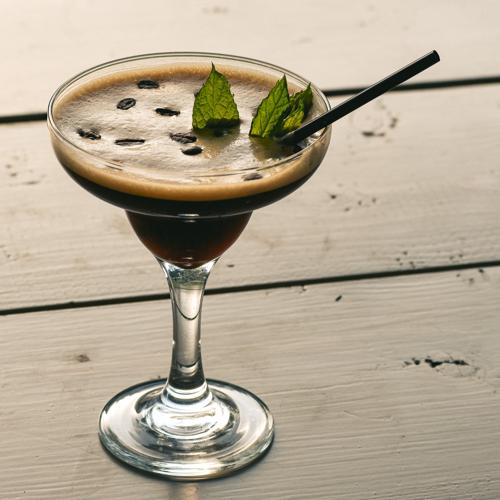} & \includegraphics[width=25mm, height=25mm]{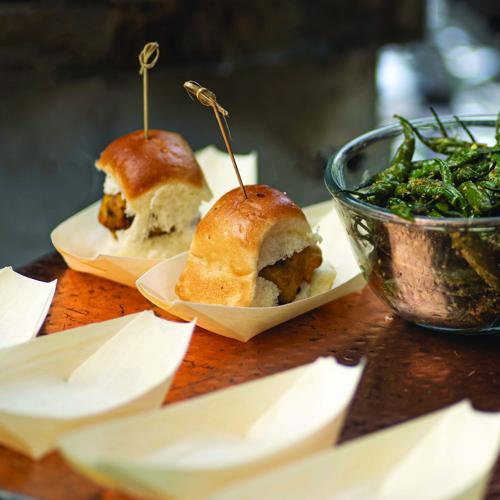} \\
 \hline
\makecell{old fashioned \\ cocktail drink} & breakfast & espresso martini & food art \\
 \hline
$\downarrow$ & kitchen & cocktail & food \\
 \hline
$\downarrow$ & $\downarrow$ & \makecell{old fashioned \\ cocktail drink} & domestic animals \\
 \hline\multicolumn{4}{|c|}{[ROOT]} \\
 \hline
\end{tabular}

\begin{tabular}{|c||c||c||c|}
 \hline
\includegraphics[width=25mm, height=25mm]{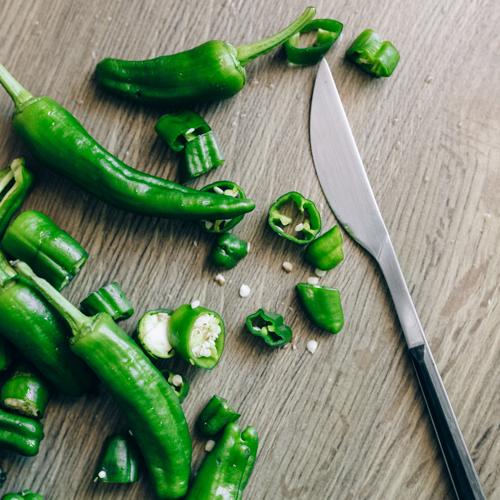} & \includegraphics[width=25mm, height=25mm]{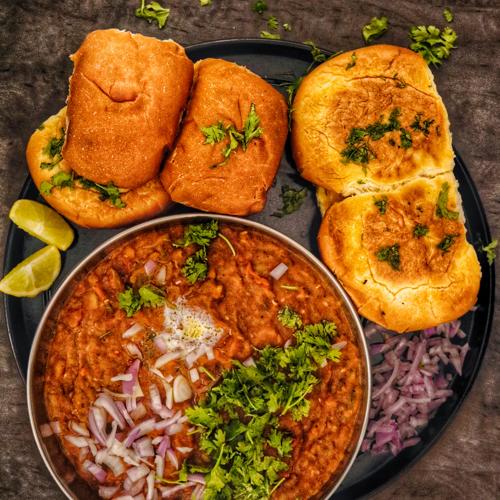} & \includegraphics[width=25mm, height=25mm]{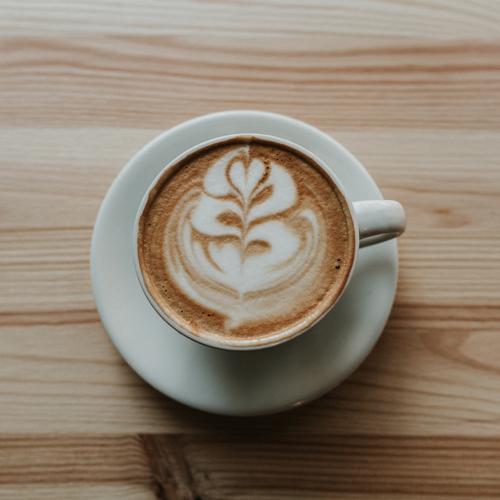} & \includegraphics[width=25mm, height=25mm]{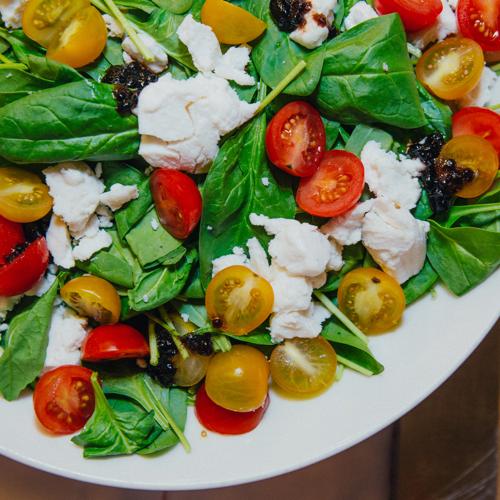} \\
 \hline
vegetable & pav bhaji & coffee & spinach caprese salad \\
 \hline
healthy eating & traditional food & $\downarrow$ & food \\
 \hline
traditional food & cityscape & $\downarrow$ & blooming flowers \\
 \hline
pets & $\downarrow$ & $\downarrow$ & kitchen \\
 \hline
domestic animal & $\downarrow$ & $\downarrow$ & $\downarrow$ \\
 \hline\multicolumn{4}{|c|}{[ROOT]} \\
 \hline
\end{tabular}

\begin{tabular}{|c||c||c||c|}
 \hline
\includegraphics[width=25mm, height=25mm]{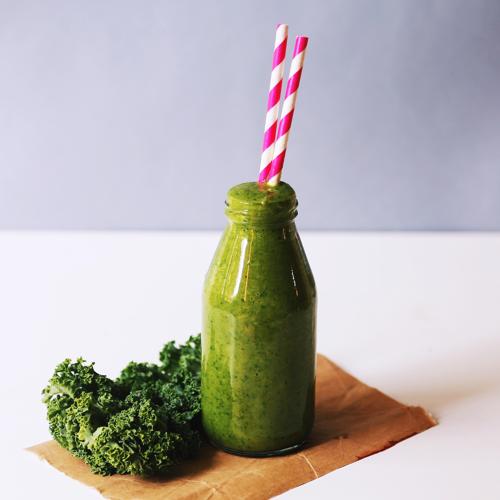} & \includegraphics[width=25mm, height=25mm]{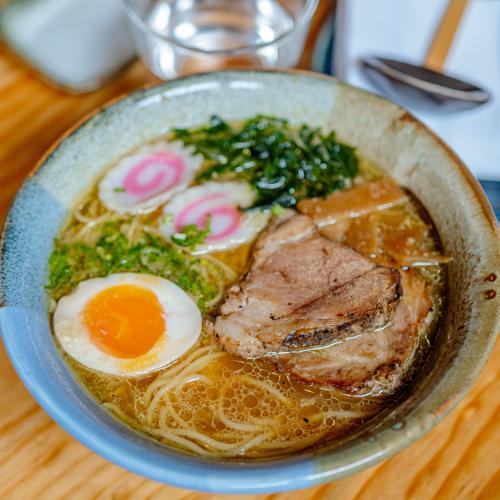} & \includegraphics[width=25mm, height=25mm]{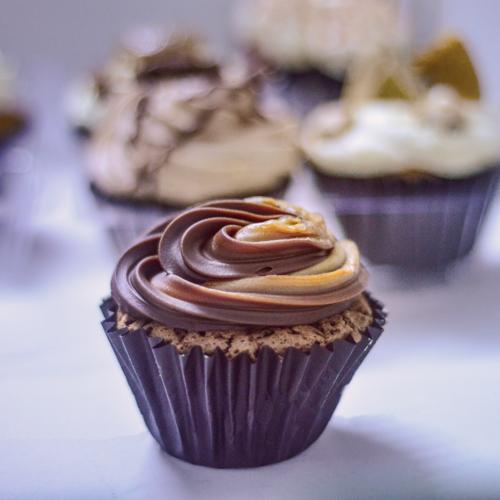} & \includegraphics[width=25mm, height=25mm]{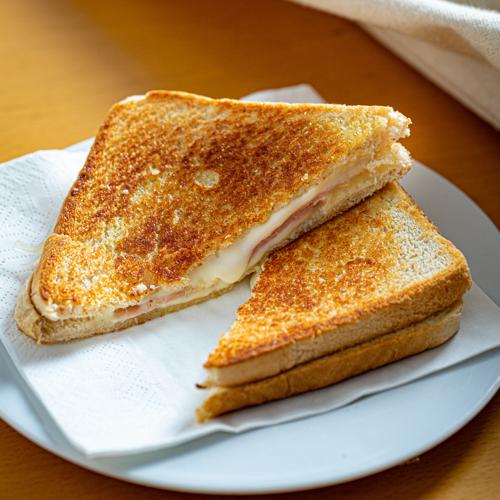} \\
 \hline
smoothie & breakfast & chocolate cupcakes & tasty \\
 \hline
beverage & tourist spot & $\downarrow$ & delicious \\
 \hline
scenery & kitchen & $\downarrow$ & traditional food \\
 \hline
kitchen & $\downarrow$ & $\downarrow$ & domestic animals \\
 \hline\multicolumn{4}{|c|}{[ROOT]} \\
 \hline
\end{tabular}

\begin{tabular}{|c||c||c||c|}
 \hline
\includegraphics[width=25mm, height=25mm]{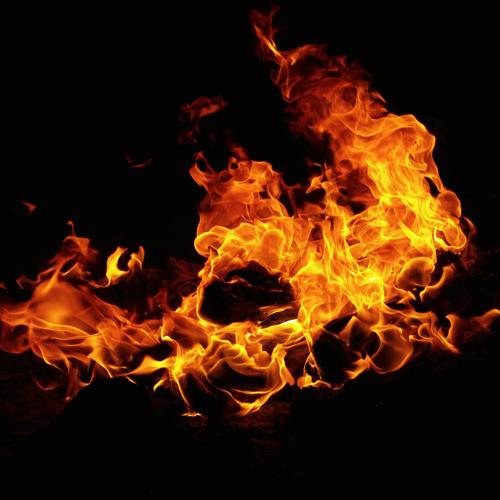} & \includegraphics[width=25mm, height=25mm]{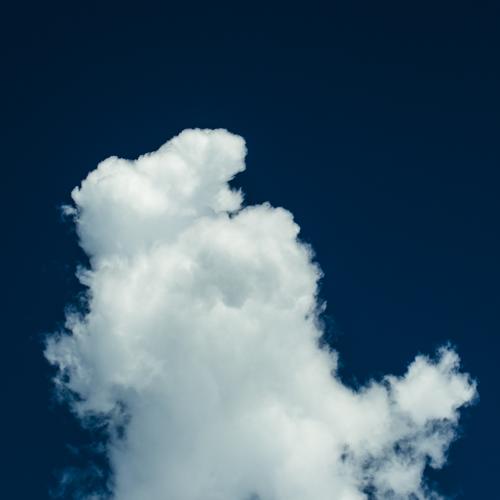} & \includegraphics[width=25mm, height=25mm]{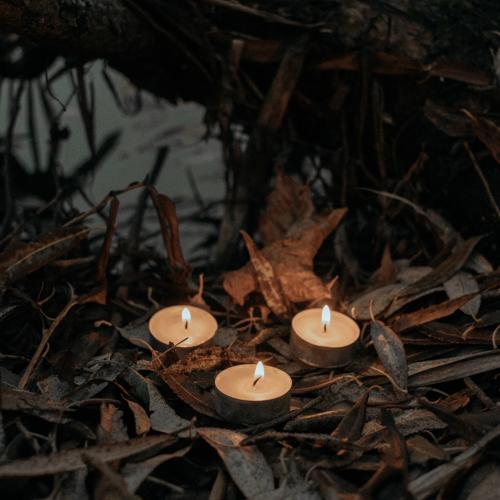} & \includegraphics[width=25mm, height=25mm]{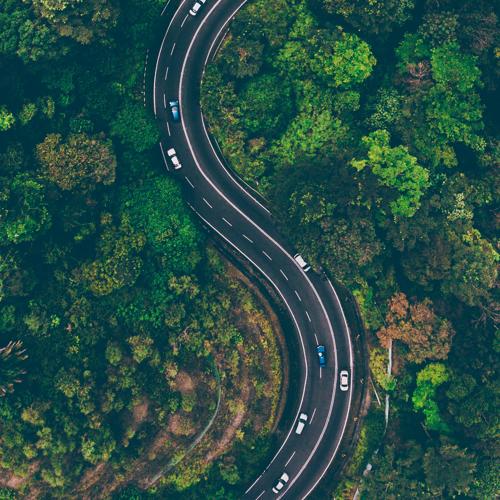} \\
 \hline
burning & cloudscape & lights & destination \\
 \hline
mountains & white clouds & outdoor & scenery \\
 \hline
$\downarrow$ & mountains & domestic animals & national park \\
 \hline
$\downarrow$ & $\downarrow$ & national park & $\downarrow$ \\
 \hline\multicolumn{4}{|c|}{[ROOT]} \\
 \hline
\end{tabular}

\begin{tabular}{|c||c||c||c|}
 \hline
\includegraphics[width=25mm, height=25mm]{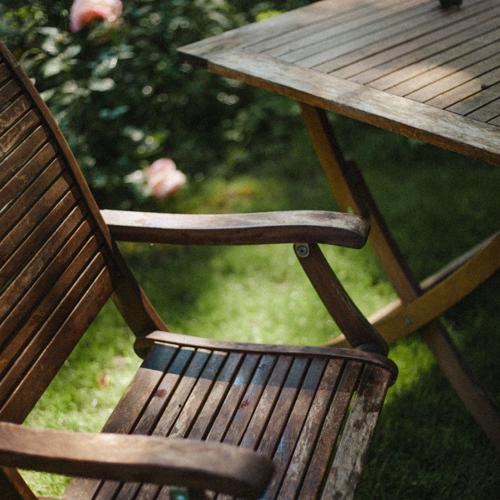} & \includegraphics[width=25mm, height=25mm]{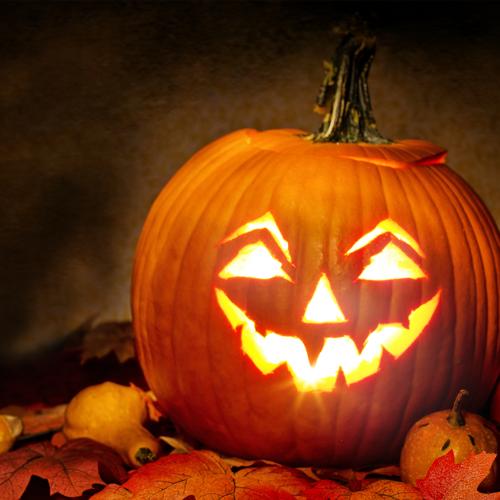} & \includegraphics[width=25mm, height=25mm]{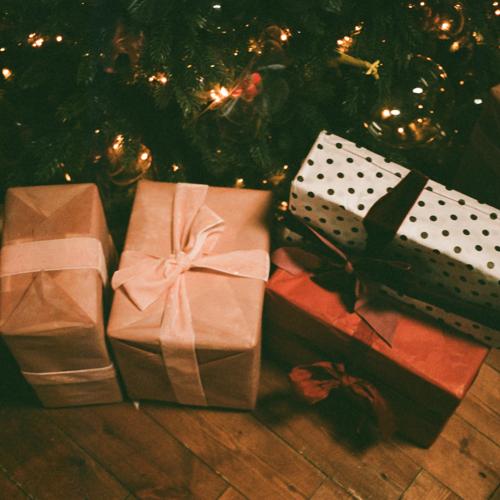} & \includegraphics[width=25mm, height=25mm]{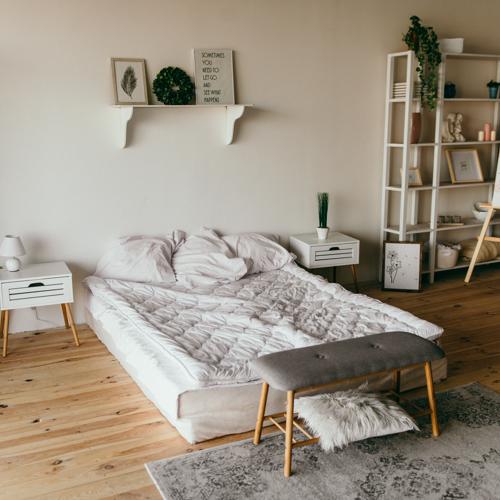} \\
 \hline
\makecell{garden table \\ and chair}  & halloween & christmas & bedroom wallpaper \\
 \hline
wooden table & domestic animals & decoration & apartment \\
 \hline
scenery & $\downarrow$ & pets & $\downarrow$ \\
 \hline
domestic animals & $\downarrow$ & domestic animals & $\downarrow$ \\
 \hline
national park & $\downarrow$ & $\downarrow$ & $\downarrow$ \\
 \hline\multicolumn{4}{|c|}{[ROOT]} \\
 \hline
\end{tabular}

\begin{tabular}{|c||c||c||c|}
 \hline
\includegraphics[width=25mm, height=25mm]{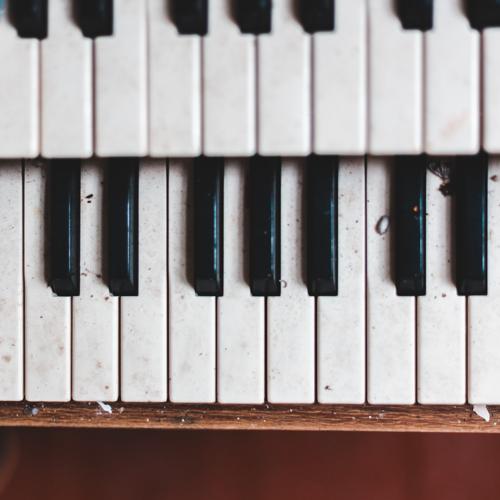} & \includegraphics[width=25mm, height=25mm]{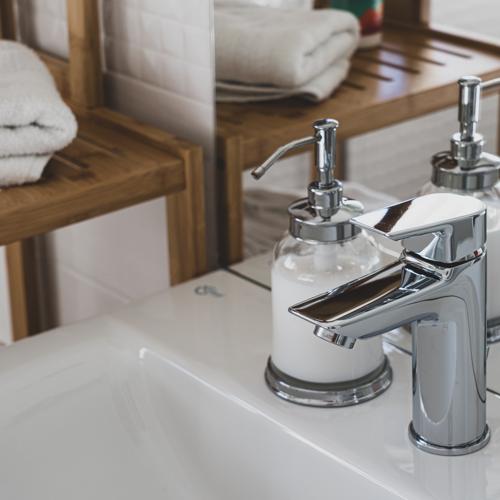} & \includegraphics[width=25mm, height=25mm]{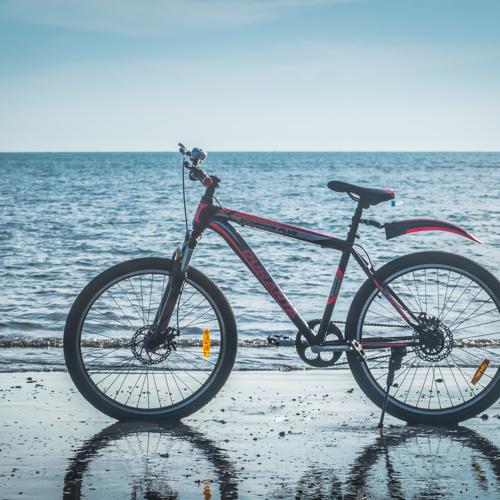} & \includegraphics[width=25mm, height=25mm]{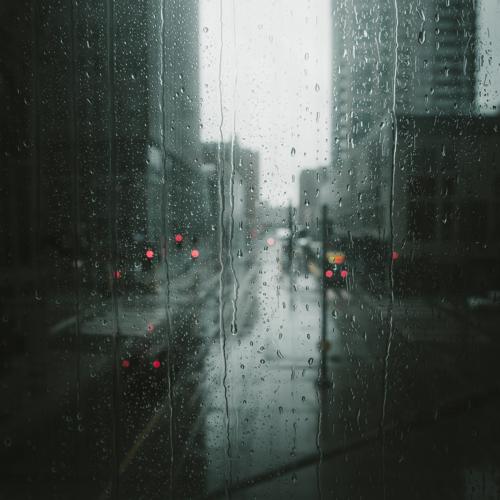} \\
 \hline
cabinet & faucet & \makecell{mountain bike \\ on the beach} & raining in the city \\
 \hline
domestic animals & \makecell{stainless steel faucet \\ on white ceramic sink} & mountain bike & new york \\
 \hline
cityscape & clean bathroom & coast & urban \\
 \hline
$\downarrow$ & bathroom & national park & city \\
 \hline
$\downarrow$ & kitchen & $\downarrow$ & street \\
 \hline
$\downarrow$ & $\downarrow$ & $\downarrow$ & cityscape \\
 \hline\multicolumn{4}{|c|}{[ROOT]} \\
 \hline
\end{tabular}

\begin{tabular}{|c||c||c||c|}
 \hline
\includegraphics[width=25mm, height=25mm]{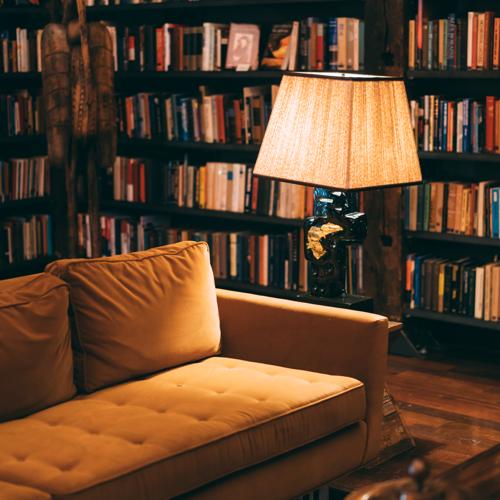} & \includegraphics[width=25mm, height=25mm]{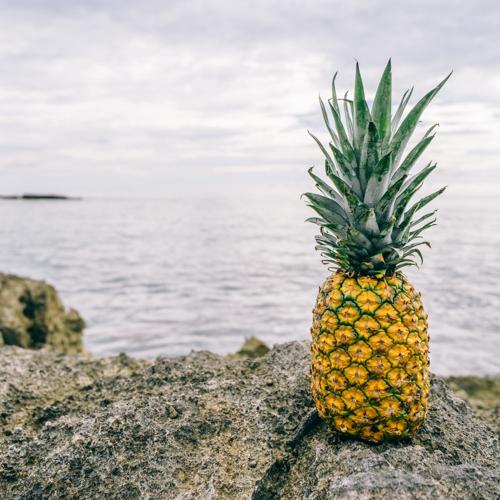} & \includegraphics[width=25mm, height=25mm]{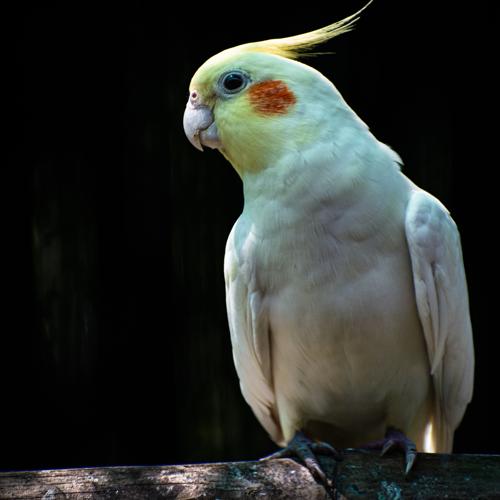} & \includegraphics[width=25mm, height=25mm]{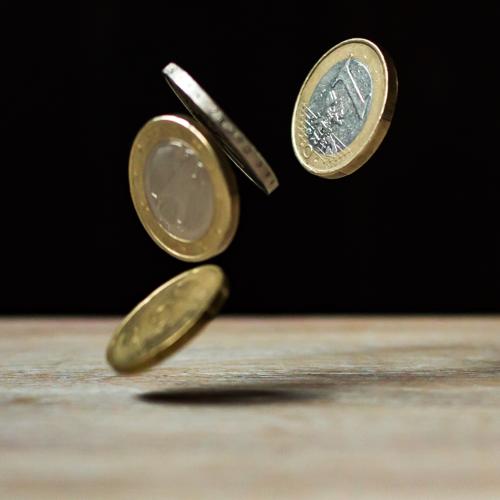} \\
 \hline
bookshelves & sea & \makecell{close-up shot \\ of a cockatiel} & antique \\
 \hline
interior design & beach & cockatiel & finance \\
 \hline
domestic animals & coastline & beak & motion \\
 \hline
cityscape & national park & animal & pets \\
 \hline
$\downarrow$ & $\downarrow$ & domestic animal & domestic animals \\
 \hline
$\downarrow$ & $\downarrow$ & $\downarrow$ & bathroom \\
 \hline\multicolumn{4}{|c|}{[ROOT]} \\
 \hline
\end{tabular}

\section{Zero-shot performance for small scale ViT-B/32 models}
\label{sec:Small-ViT-B-32-Models}

\begin{table}[h!]
\caption{Zero-shot performance for small scale ViT-B/32 models.}
\label{table:Small-ViT-B-32-Models}
\centering
\begin{tabular}{|c|c|c|c|c|c|c|} 
 \hline
  & & & ImageNet &  &  & Average over \\
  \multirow{-2}{*}{Geometry} & \multirow{-2}{*}{Variant} & \multirow{-2}{*}{ImageNet} & dist. shifts & \multirow{-2}{*}{VTAB} & \multirow{-2}{*}{Retrieval} & 38 datasets \\
 \hline\hline
 \multicolumn{2}{|c|}{CLIP} & \underline{4.96} & \textbf{5.5} & \underline{17.2} & \underline{11.4} & 16.2\\
 \hline
 \multicolumn{2}{|c|}{Elliptic} & \textbf{4.99} & \underline{5.4} & \textbf{17.3} & \textbf{11.5} & 15.9\\
 \hline
  &  EuCLIP  & 3.476 & 4.3& 15.2& 10.7& 14.4\\
 \cline{2-7}
  & $d^2$, no-ln, $\lambda$=0 & 4.18 & 4.9& 15.9& 10.9& 15.0\\
 \cline{2-7}
 \multirow{-3}{*}{Euclidean} & $d$, ln, $\lambda$=0 & 4.54 & 5.3& 16.9& 11.3& 15.9\\
 \hline
  & $d^2$, no-ln, $\lambda$=0.2 & 2.354 & 3.6& 14.3& 10.1& 13.4\\
 \cline{2-7}
  & $d^2$, no-ln, $\lambda$=0 & 4.05 & 4.8& 16.1& 10.9& 15.0\\
 \cline{2-7}
  & MERU & 3.214 & 4.4& 15.5& 10.5& 14.2\\
 \cline{2-7}
 \multirow{-4}{*}{Hyperbolic} & $d$, ln, $\lambda$=0 & 4.18 & 5.0& 16.5& 11.2& 15.7\\
 \hline
\end{tabular}
\end{table}

\newpage

\section{Embedding Distance Distributions}
\label{sec:embedding_dist_dist}
\begin{figure}
    \centering
    \includegraphics[width=0.75\linewidth]{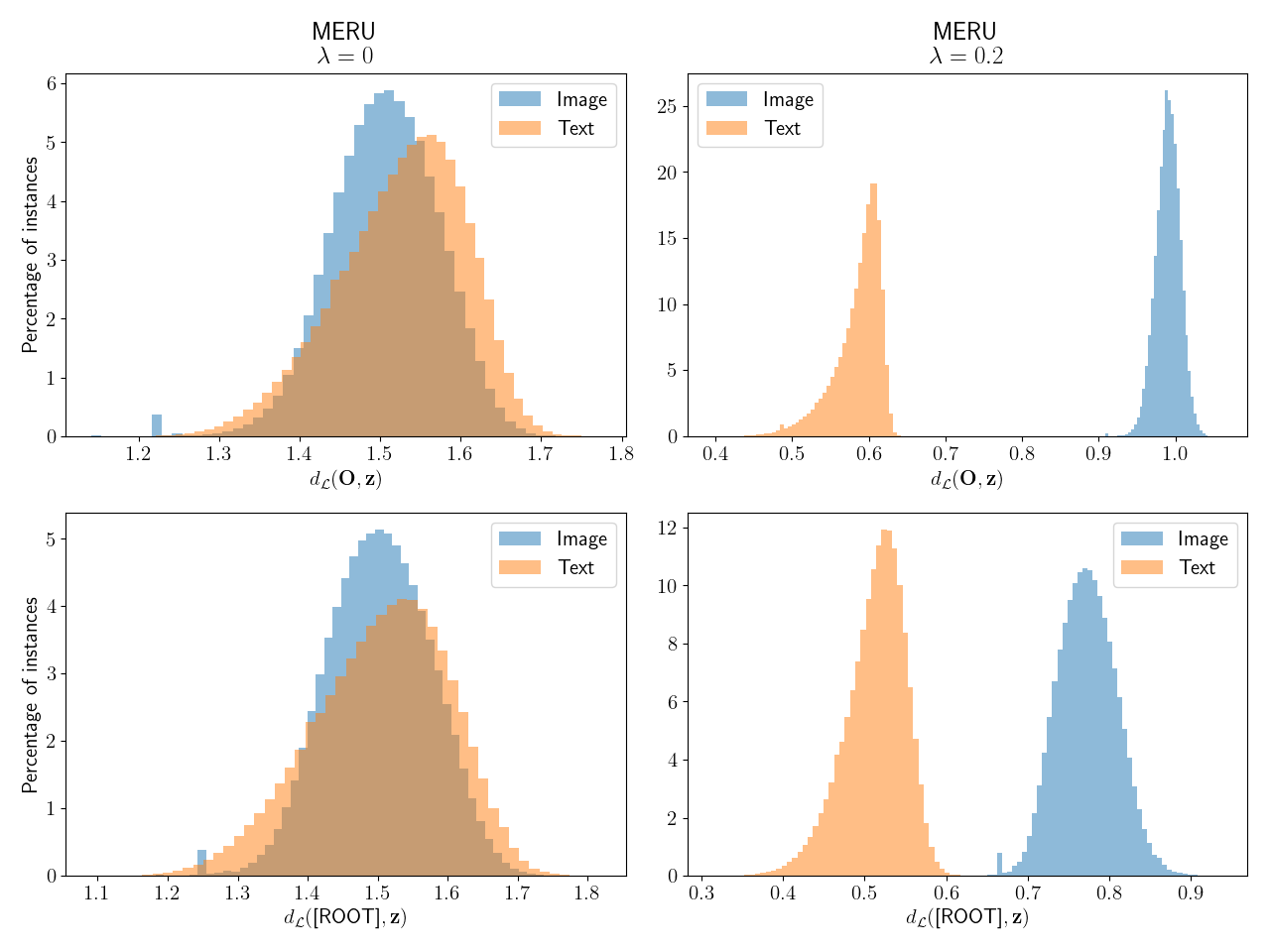}
    \caption{Distribution of embedding distances from the origin $\mathbf{O}$ (upper) vs. from the embedding average [ROOT] (lower) for ViT-B/32 MERU Models, $\lambda = 0$ (left) vs. $\lambda = 0.2$ (right). Unexpectedly, the embedding average [ROOT] deviates further from the origin $\mathbf{O}$ with entailment loss, indicating asymmetric distribution.}
\label{fig:zero_vs_average_root_dist_distribution}
\end{figure}

\begin{figure}
    \centering
    \includegraphics[width=0.75\linewidth]{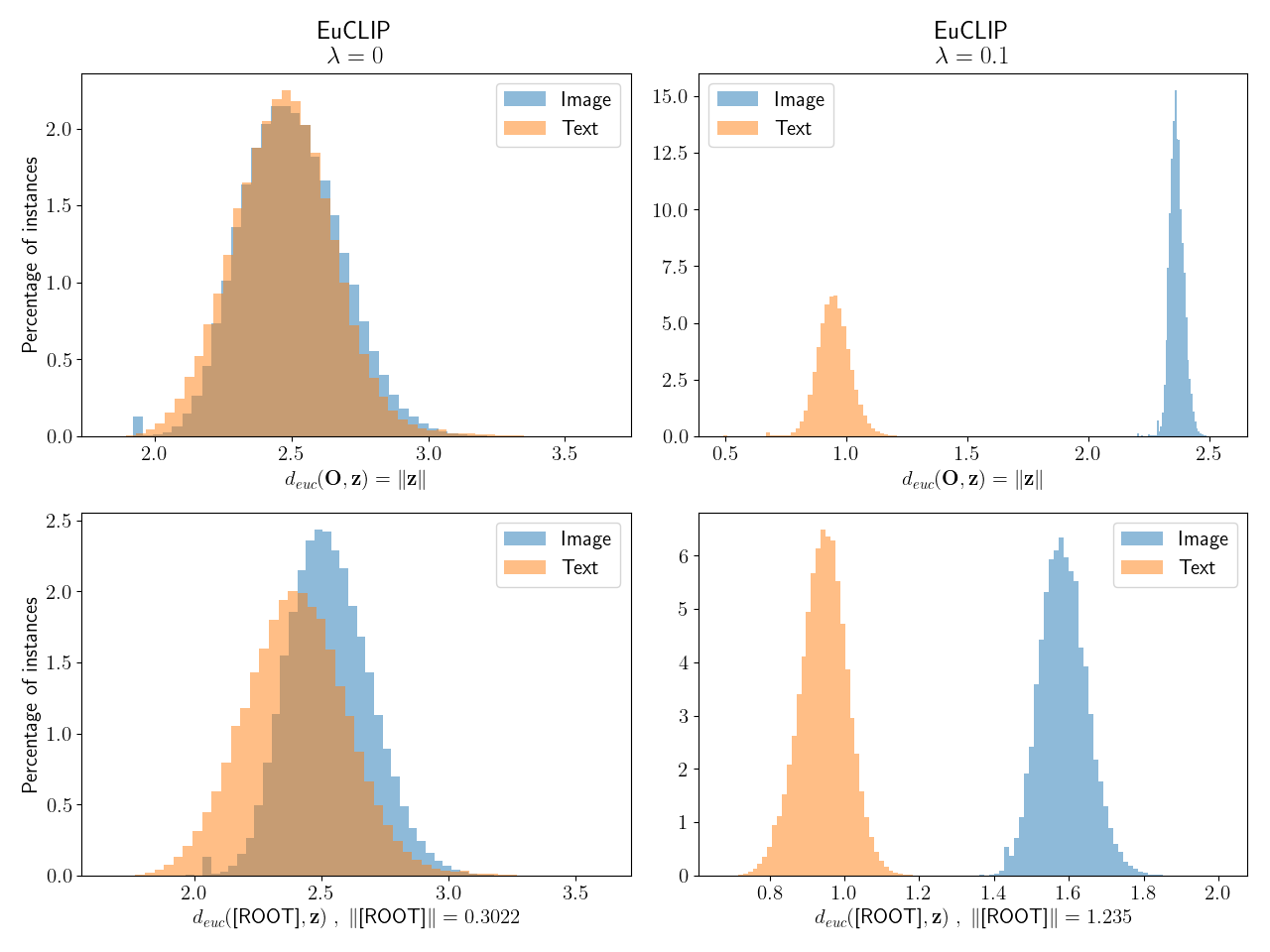}
    \caption{Distribution of embedding distances from the origin $\mathbf{O}$ (upper) vs. from the embedding average [ROOT] (lower) for ViT-B/32 EuCLIP Models, $\lambda = 0$ (left, $\|$[ROOT]$\| = 0.3022$) vs. $\lambda = 0.1$ (right, $\|$[ROOT]$\| = 1.235$).}
    \label{fig:euclip_zero_vs_average_root_dist_distribution}
\end{figure}

\begin{figure}
    \centering
    \includegraphics[width=0.75\linewidth]{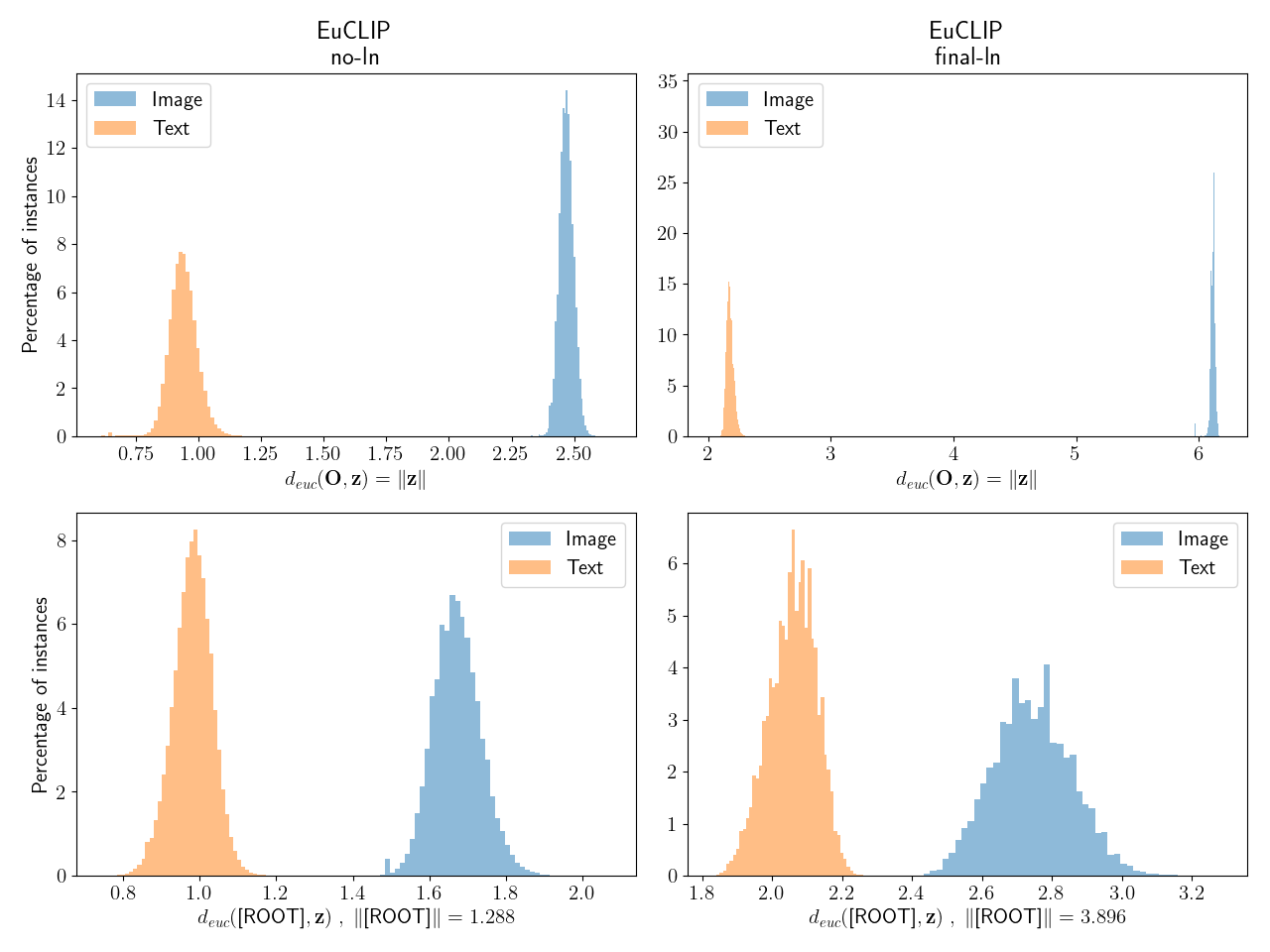}
    \caption{Distribution of embedding distances from the origin $\mathbf{O}$ (upper) vs. from the embedding average [ROOT] (lower) for ViT-B/16 EuCLIP Models, no-ln (left, $\|$[ROOT]$\| = 1.288$) vs. final-ln (right, $\|$[ROOT]$\| = 3.896$).}
    \label{fig:euclip_ln_zero_vs_average_root_dist_distribution}
\end{figure}

For both MERU (Figure \ref{fig:zero_vs_average_root_dist_distribution}) and EuCLIP, the average embedding deviates further from the origin $\mathbf{O}$ in models trained with entailment loss than without. For example, the L2 norm of [ROOT] for the medium scale ViT-B/32 EuCLIP $\lambda = 0$ model is 0.3022 but that of the $\lambda = 0.1$ model is 1.235 (Figure \ref{fig:euclip_zero_vs_average_root_dist_distribution}). For the ViT-B/16 EuCLIP models, the L2 norm of [ROOT] of the $\lambda = 0.1$ model is 1.288, almost the same as that of the ViT-B/32 counterpart, but that of the $\lambda = 0.1$ model with the final LN is 3.896, unexpectedly large in terms of magnitude (Figure \ref{fig:euclip_ln_zero_vs_average_root_dist_distribution}). We understand and expect poor performance of the final LN model due to loss of the degree of freedom, but we do not understand how it results in such off-center embedding distribution in combination with the entailment loss.

\end{document}